\documentclass{article}

\PassOptionsToPackage{numbers, compress}{natbib}

 \usepackage[main, final]{neurips_2026}

\usepackage{xurl}
\usepackage[utf8]{inputenc} 
\usepackage[T1]{fontenc}    
\usepackage{hyperref}       
\usepackage{url}            
\usepackage{booktabs}       
\usepackage{amsfonts}       
\usepackage{nicefrac}       
\usepackage{microtype}      
\usepackage{xcolor}         

\usepackage{graphicx}
\usepackage{enumitem}
\usepackage{multirow}
\usepackage{booktabs}
\usepackage[table]{xcolor}
\usepackage{graphicx}
\usepackage{amsmath}
\usepackage{wrapfig}

\usepackage{algorithm}
\usepackage[most]{tcolorbox}

\usepackage{amsmath}
\usepackage{amssymb}
\usepackage{bm}

\usepackage{algorithmic}

\title{PhGPO: Pheromone-Guided Policy Optimization for Long-Horizon Tool Planning}

\author{%
  Yu Li\textsuperscript{1},
  Guangfeng Cai\textsuperscript{1},
  Shengtian Yang\textsuperscript{1},
  Han Luo\textsuperscript{1}, 
  Shuo Han\textsuperscript{2},
  Xu He\textsuperscript{2},
  Dong Li\textsuperscript{2},
  Lei Feng\textsuperscript{1}\thanks{Corresponding author.} \\[0.5em]
  \textsuperscript{1}School of Computer Science and Engineering \\
  Southeast University, Nanjing, China \\
  \textsuperscript{2}Huawei Noah's Ark Lab, China \\
\texttt{\{yuli11,fenglei\}@seu.edu.cn}
}

\begin{document}

\maketitle

\begin{abstract}
Recent advancements in Large Language Model (LLM) agents have demonstrated strong capabilities in executing complex tasks through tool use. However, long-horizon multi-step tool planning is challenging, because the exploration space suffers from a combinatorial explosion. In this scenario, even when a correct tool-use path is found, it is usually considered an immediate reward for current training, which would not provide any reusable information for subsequent training.
In this paper, we argue that \emph{historically successful trajectories contain reusable tool-transition patterns}, which can be leveraged throughout the whole training process.
Inspired by ant colony optimization where historically successful paths can be reflected by the \emph{pheromone}, we propose Pheromone-Guided Policy Optimization (PhGPO), which learns a trajectory-based transition pattern (i.e., pheromone) from historical trajectories and then uses the learned pheromone to guide policy optimization.
This learned pheromone provides explicit and reusable guidance that steers policy optimization toward historically successful tool transitions, thereby improving long-horizon tool planning.
\end{abstract}

\section{Introduction}

Large Language Model (LLM) agents have demonstrated strong capabilities in executing complex tasks through tool use \citep{AgentOrchestra,MCP-Universe,toolformer, yang, li2026learning}. By invoking external tools such as web search \citep{WebArena,phase}, computer operation \citep{Alita}, and code execution \citep{code}, agents can access up-to-date information and take actions beyond static parametric knowledge. 
As the tool space expands, many real-world tasks require a tool-use trajectory, where early choices constrain later reachable states and determine whether the entire trajectory succeeds \citep{react,Tool-Planner}.
When tool-use trajectories extend to ten to twenty steps, as in Toolathlon \citep{toolathlon}, long-horizon tool planning becomes difficult: agents must choose from hundreds of tools, yet even when a successful trajectory is discovered, its transition patterns are hard to explicitly store and reliably reuse, which hinders effective learning in such a large tool space \citep{toucan,traject}.

Reinforcement learning (RL) offers a natural way to leverage interaction experiences and feedback to update the policy for long-horizon tool planning \citep{steptool,torl,treerl}. However, interaction experiences are typically absorbed implicitly into the policy parameters, leaving no explicit record of historically successful tool transitions.
As a result, even after a successful tool-use trajectory is discovered, its transition patterns remain implicit in the policy parameters and cannot be directly reused as guiding information. When successful trajectories are rare, the exploration of RL becomes inefficient because the agent must search again for similar long-horizon trajectories \citep{arpo,salt}. 

\begin{figure*}[t]
  \begin{center}
    \centerline{\includegraphics[width=\textwidth]{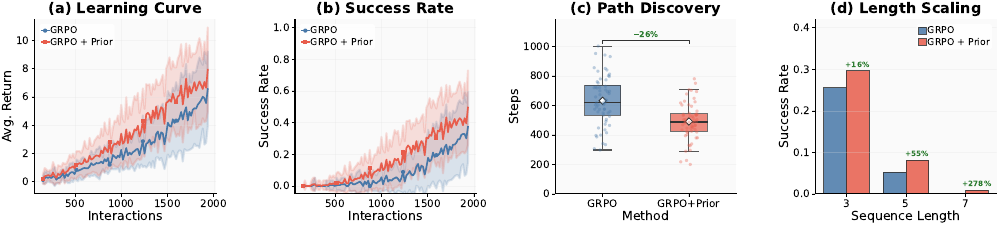}}
    \caption{\textbf{Two-stage pilot experiment with an explicit transition prior for long-horizon tool planning.} We distill cross-trajectory transition memory from verified successful tool-use trajectories in an easier stage, then reuse it as a prior for GRPO optimization. Panels (a,b) show higher average return and success rate under matched interaction budgets, (c) fewer steps to the first successful tool-use trajectory, and (d) larger relative gains with longer trajectories.}
    \label{intro}
  \end{center}
  \vskip -0.8cm
\end{figure*}

In this paper, we argue that \emph{an explicit transition prior} can be distilled from historically successful trajectories, which would be helpful to policy optimization. We validate our viewpoint with a two-stage pilot experiment (Figure~\ref{intro}): Firstly, we collect verified successful tool-use trajectories and distill their transition patterns into an explicit transition prior. Secondly, we use this prior to reweight the policy’s next-tool distribution in accord with verified transitions. 
As shown in Figure~\ref{intro}, the explicit transition prior improves both sample efficiency and final success, 
with larger gains as trajectory length increases.

Motivated by the above analysis, we would like to seek an explicit transition prior that can be learned from historical trajectories and reused during policy optimization. To realize such a prior, we draw inspiration from Ant Colony Optimization (ACO), where the \emph{pheromone} summarizes historically successful paths and encourages them to be taken again \citep{aco,pheromone}. 
Based on this inspiration, we propose \textbf{Pheromone-Guided Policy Optimization (PhGPO)}, which learns an explicit transition prior from historical trajectories, and this transition prior acts as the pheromone in ACO. Concretely, PhGPO organizes tools into a tool-transition graph based on the Model Context Protocol (MCP) \citep{MCP-Universe} and associates each transition with a pheromone value. During policy optimization, we integrate the pheromone values into next-step tool selection, making historically successful tool transitions more likely to be reproduced. 
It is noteworthy that the pheromone is reinforced by verified successful tool-use trajectories and decays over time, which keeps the transition pattern reusable and up to date.
We further construct a set of frequent argument invocations for each tool, enabling the agents to learn not only which tool to call but also how to invoke it effectively. 

Based on the pheromone mechanism as described above, our PhGPO adopts a progressive training pipeline to equip an agent with long-horizon tool planning abilities. Firstly, PhGPO begins with supervised warm-up using a next-tool prediction objective to provide a stable initialization. 
Secondly, PhGPO performs pheromone-guided reinforcement learning with a progressive schedule: it injects the correct next tool with a decaying probability and progressively cedes control to the policy, which enables pheromone to be continually updated from accumulated policy-driven successes.  As the accumulated pheromone becomes more reliable, it fully leverages pheromone guidance for trajectory generation to optimize long-horizon performance.
We empirically evaluate PhGPO on three long-horizon tool-planning benchmarks, including Toolathlon \citep{toolathlon}, TOUCAN \citep{toucan} and TRAJECT-Bench \citep{traject}, which cover diverse real-world tool-use scenarios with long multi-step tool-use trajectories. Across these benchmarks, PhGPO consistently produces tool-use trajectories that more closely match the reference trajectories on complex tasks, demonstrating that pheromone enables effective reuse of historical successes and improves long-horizon tool planning performance.
Our main contributions can be summarized as follows:
\begin{itemize}[leftmargin=0.5cm,topsep=0pt]
\item \textbf{Explicit reuse in long-horizon tool planning.} We show that transition patterns in successful tool-use trajectories are typically absorbed implicitly into the policy parameters, making them hard to reuse and slowing learning when verified successes are rare.
\item \textbf{PhGPO with pheromone as an explicit transition prior.} We propose PhGPO, which distills successful trajectories into an explicit transition prior working as pheromone, and uses it to guide policy optimization toward historically successful tool transitions.
\item \textbf{MCP-grounded graph and progressive training.} We build an MCP-grounded tool-transition graph to maintain pheromone and introduce a progressive training pipeline to accumulate and exploit it over long tool-use trajectories, achieving consistent gains on Toolathlon, 
TOUCAN, and TRAJECT-Bench.

\end{itemize}

\section{Related Work}

\noindent\textbf{Retrieval-augmented tool use.}
Tool selection at scale is a central challenge for tool-using agents. Many methods use retrieval to narrow the candidate set by fetching relevant documentation, schemas, or demonstrations \citep{react,reflexion, min, min2}. Early prompting-based agents interleave reasoning and action selection over retrieved evidence \citep{react,reflexion}, while recent systems further train retrievers or couple retrieval with tool-calling models to improve robustness under large and evolving tool pools \citep{gorilla,toolformer,toollLM,stabletoolbench,apibank}. Dedicated benchmarks and pipelines also treat tool selection as an information retrieval problem \citep{toolret,toolreagt,toolgen,online}. While retrieval improves per-step grounding, long-horizon tool-use trajectories still require consistent transition patterns across multiple decisions.

\noindent\textbf{Graph-structured tool planning and search.}
Beyond retrieval, many methods organize APIs and dependencies into graphs or trees, and then perform graph traversal, heuristic search, or learned graph-based planning \citep{toolnet,controlllm,naviagent,gtool,toolchain}. Some approaches search or expand over tool graphs to produce multi-step tool-use trajectories \citep{toolnet,controlllm,naviagent}, while others generate executable tool graphs or DAG-like plans \citep{gtool}. Interactive search and self-updating mechanisms can also improve robustness to evolving tool interfaces \citep{learningevolvingtools}. These structured methods reduce the effective search space, but they usually rely on heuristic or supervised planning signals and do not maintain an adaptive transition prior distilled from accumulated successful trajectories.

\noindent\textbf{Reinforcement learning for tool use.}
Reinforcement learning has been increasingly used to improve multi-step tool use under sparse outcomes \citep{steptool,torl}. Recent work studies tool-integrated optimization with verifiable or execution-based rewards \citep{autotir}, hierarchical agent designs that separate high-level planning from tool execution \citep{agent_as_tool}, and executable-code agents that learn tool-use strategies from outcomes \citep{retool,toolstar}. Despite these advances, successful experience is often absorbed implicitly into policy parameters, which can slow learning when verified successful trajectories are rare.

\noindent\textbf{Experience reuse and agent memory.}
Explicit experience reuse stores and replays successful behaviors instead of relying only on implicit parameter updates \citep{agentkb,aflow}. Record-and-replay paradigms cache trajectories or routines to accelerate future problem solving \citep{agentdistill,get_experience_from_practice}. However, these methods rarely convert accumulated successes into structured transition statistics that can serve as an explicit transition prior for long-horizon tool-use trajectories.

\noindent\textbf{Ant-inspired reinforcement learning and pheromone-based search.}
Prior ant-inspired methods use pheromone mechanisms for search, routing, multi-agent coordination, and optimal control \citep{dorigo1996ant,dorigo2002ant,gambardella1995antq,dicaro1998antnet,monekosso2004pheq,vanast2010antcolony}, and recent LLM work applies similar signals to inference-time reasoning-path search \citep{chari2025pheromone}. PhGPO differs by treating pheromone as an explicit transition prior for long-horizon tool-use policy optimization, not as a stand-alone optimizer or value-like controller. This prior is distilled from verified successful tool-use trajectories and used as a log-space prior during rollout generation, while policy learning remains governed by GRPO.

\section{Methodology}
\label{sec:method}



In this section, we present Pheromone-Guided Policy Optimization (\textbf{PhGPO}) for long-horizon tool planning. PhGPO is built around a simple principle: long-horizon tool use should explicitly \emph{accumulate} and \emph{reuse} historically successful tool transitions from verified tool-use trajectories. Inspired by \emph{Ant Colony Optimization (ACO)} \citep{aco}, PhGPO instantiates the explicit transition prior as pheromone on a tool-transition graph. Pheromone is reinforced by verified successful trajectories and decays over time, yielding an adaptive supervision signal that can be incorporated into next-step tool selection for policy optimization.
In this way, PhGPO converts trajectory-level success into transition-level pheromone on the graph, which steers policy optimization toward historically successful tool transitions while avoiding stale bias through decay.

\begin{figure*}[t]
  \begin{center}
    \centerline{\includegraphics[width=\textwidth]{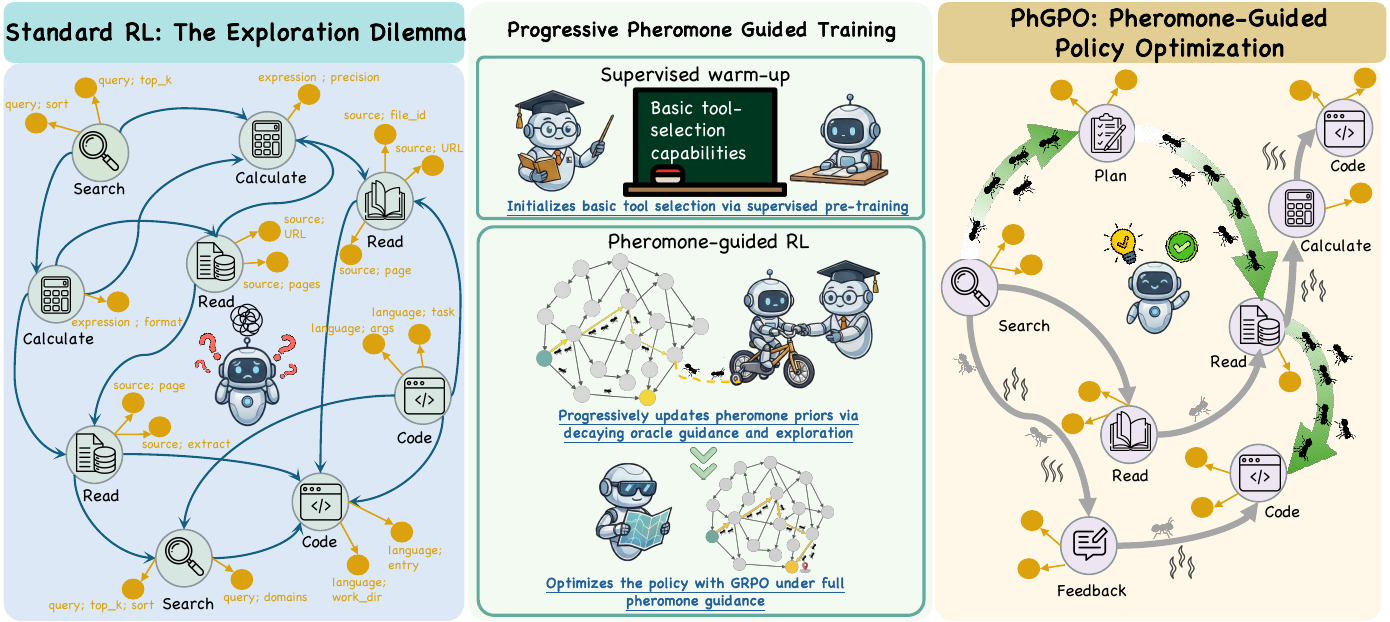}}
    \caption{\textbf{Overview of PhGPO.} PhGPO converts verified successful tool-use trajectories into a pheromone-based explicit transition prior over tool-transition and tool-to-invocation edges. The prior is updated through deposition and evaporation, and incorporated into rollout generation to guide the policy toward historically successful transitions and argument invocations. Training proceeds from supervised next-tool warm-up to progressive pheromone-guided reinforcement learning and finally full pheromone-guided policy optimization.}
    \label{method}
    
  \end{center}
  \vskip -0.6cm
\end{figure*}

\subsection{Tool-Transition Graph from MCP}
\label{sec:tool-graph}
The goal of long-horizon tool use is to first select the next tool and then specify an executable invocation. To this end, we build a tool-transition graph from MCP that captures tool transitions from verified trajectories and explicitly represents tools and their transitions, providing a structured space for modeling multi-step tool-use trajectories and their transition patterns.  We also construct a set of frequent argument invocations for each tool, which captures how tool arguments are frequently invoked.

\noindent\textbf{Tool-transition graph.}
Let $\mathcal{V}_{\text{Tool}}=\{v_1,\dots,v_N\}$ be the base tools exposed by MCP.
We view a tool-use trajectory as a path over $\mathcal{V}_{\text{Tool}}$ and define a directed edge $(v_i,v_j)\in\mathcal{E}$ when $v_j$ follows $v_i$ in a trajectory.
This graph captures tool-to-tool dependencies and provides the substrate for accumulating transition experience on edges.

\noindent\textbf{Argument-invocation set.}
The invocation of each tool $v_i\in\mathcal{V}_{\text{Tool}}$ also depends on the specific structure of tool arguments.
For each tool $v_i\in\mathcal{V}_{\text{Tool}}$, 
we construct a set of frequent executable argument invocations $\mathcal{V}_{\text{Arg}}$ that captures how tool arguments are frequently invoked, denoted by $\mathcal{V}_{\text{Arg}}(v_i)=\{v_i^{(1)},\dots,v_i^{(K_i)}\}$ where $v_i^{(j)}$ denotes the $j$-th argument-invocation pattern among $K_i$ key patterns of the tool $v_i$.
To make the mapping explicit, we define tool-to-invocation association edges
$\mathcal{E}_{\text{arg}}=\bigcup_{i=1}^{N}\{(v_i, v_i^{(k)}) \mid v_i^{(k)}\in\mathcal{V}_{\text{Arg}}\}$.
An edge $(v_i, v_i^{(k)})\in\mathcal{E}_{\text{arg}}$ indicates that $v_i^{(k)}$ is an argument-invocation associated with tool $v_i$.

Based on both the tool-transition graph and the argument-invocation set, we can have a factorized decision at step $t$: the agent first selects the next tool $a_t\in\mathcal{V}_{\text{Tool}}$, then selects an argument invocation $u_t\in\mathcal{V}_{\text{Arg}}(a_t)$ to instantiate an executable invocation template.

\subsection{Pheromone as an Explicit Transition Prior}
\label{sec:pheromone}
Our PhGPO integrates both \emph{task-agnostic} pheromones with \emph{task-dependent} pheromones from verified successful tool-use trajectories.

\noindent\textbf{Task-agnostic pheromones.}
We maintain two task-agnostic pheromones, including
$\tau^{\text{agn}}_{\text{Tool}}(v_i,v_j)$ over tool-transition edges
and $\tau^{\text{agn}}_{\text{Arg}}(v_i,v_i^{(k)})$ over tool-to-invocation edges.
After executing a trajectory $\xi$, we compute a bounded pheromone-deposition score
$q_{\tau}(\xi)\in[0,1]$, measured by the match ratio to the reference tool-use trajectory.
This score is used only to determine how much pheromone is deposited on the traversed edges.
It is not used as a direct step-level reward and does not directly enter the policy optimization loss.
For each traversed tool transition and each selected argument invocation in $\xi$, we apply ACO-style deposition and evaporation:
\begin{gather}
\tau^{\text{agn}}
\leftarrow
\mathrm{clip} \bigl((1-\rho)\tau^{\text{agn}}+\alpha\,q_{\tau}(\xi),~[\tau_{\min},\tau_{\max}]\bigr).
\label{eq:agn}
\end{gather}
Here, $\rho\in(0,1)$ is the evaporation rate, $\alpha>0$ is the deposition rate, and
$q_{\tau}(\xi)$ scales the amount of pheromone deposited by trajectory $\xi$.
Higher-quality tool-use trajectories contribute stronger pheromone updates, while evaporation reduces the influence of older statistics.
We clip pheromones to $[\tau_{\min},\tau_{\max}]$ to prevent excessive concentration on a small subset of transitions.
It is noteworthy that the construction of both $\tau^{\text{agn}}_{\text{Tool}}$ and $\tau^{\text{agn}}_{\text{Arg}}$ follows the same way as shown in Eq.~(\ref{eq:agn}).

\noindent\textbf{Task-dependent pheromones.}
For each transition $(v_i,v_j)$, we maintain an edge memory bank $\mathcal{M}_{ij}$ that stores tuples $(\bm{e},q_{\tau})$, where $\bm{e}$ is a task embedding and $q_{\tau}$ is the pheromone-deposition score of a verified successful tool-use trajectory that includes this transition.
Given a task description $x$, we compute $\bm{e}_x=\phi(x)$ using a sentence encoder $\phi$ and retrieve memories whose cosine similarity exceeds a threshold:
\begin{equation}
\mathcal{M}_{ij}(x)=\{(\bm{e},q_{\tau})\in\mathcal{M}_{ij}\mid \mathrm{sim}(\bm{e}_x,\bm{e})\ge \theta_{\text{sim}}, \forall \bm{e}\}.
\end{equation}
Then, we estimate task-dependent pheromone by similarity-weighted averaging:
\begin{equation}
\tau^{\text{dep}}(v_i,v_j\mid x) = \tau_0 + 
\frac{\sum_{(\bm{e},q_{\tau})\in \mathcal{M}_{ij}(x)} 
\mathrm{sim}(\bm{e}_x,\bm{e})\,q_{\tau}}
{\sum_{(\bm{e},q_{\tau})\in \mathcal{M}_{ij}(x)} 
\mathrm{sim}(\bm{e}_x,\bm{e})+\epsilon_{\mathrm{dep}}} 
\cdot (\tau_{\max}-\tau_0).
\label{eq:dep}
\end{equation}
where $\tau_0$ is a base value, $\epsilon_{\mathrm{dep}}$ is a small constant for numerical stability, and the scaling $(\tau_{\max}-\tau_0)$ keeps $\tau^{\text{dep}}$ in a comparable range.
It is noteworthy that the construction of both $\tau^{\text{dep}}_{\text{Tool}}(v_i,v_j\mid x)$ and $\tau^{\text{dep}}_{\text{Arg}}(v_i,v_i^{(k)}\mid x)$ follows the same way as shown in Eq.~(\ref{eq:dep}).

Nevertheless, as shown in Eq.~(\ref{eq:dep}), if the edge memory bank $\mathcal{M}_{ij}$ contains very few memories $(\bm{e},q_{\tau})$, the estimated task-dependent pheromone could be unreliable. Therefore, we need to assign a corresponding confidence $c_{ij}(x) \in [0,1]$ to each task-dependent pheromone for representing their reliability:
\begin{gather}
c_{ij}(x) = \min(1, n_{\text{act}} / n_{\min}) \cdot s_{\max} \cdot \bar{q}_{\tau},
\label{eq:confidence}
\end{gather}
where $n_{\text{act}}$ is the number of memories in $\mathcal{M}_{ij}$,
$n_{\min}$ is a pre-defined minimum number of elements in $\mathcal{M}_{ij}$, $s_{\max}$ is the maximum similarity observed, and $\bar{q}_{\tau}$ is the mean pheromone-deposition score of the retrieved memories. In this way, the final pheromone can be fused by simultaneously considering task-agnostic pheromones and task-dependent pheromones:
\begin{equation}
\tau_{\text{Tool}}(v_i,v_j\mid x) = (1 - w \cdot c_{ij}(x)) \cdot 
\tau^{\text{agn}}_{\text{Tool}}(v_i,v_j) + w \cdot c_{ij}(x) \cdot \tau^{\text{dep}}_{\text{Tool}}(v_i,v_j\mid x).
\label{eq:fuse-pher}
\end{equation}
where $w$ controls the maximum contribution of task dependence. We apply the same fusion way as shown in Eq.~(\ref{eq:fuse-pher}) for calculating $\tau_{\text{Arg}}$.

\subsection{Pheromone-Guided Policy Optimization Training}
\label{sec:aco-grpo}
\label{sec:training}
Learning long-horizon tool planning from scratch is difficult because early training lacks reliable successes to build an explicit transition prior. We adopt a progressive pipeline that stabilizes basic behavior, gradually accumulates pheromone from verified tool-use trajectories, and then performs pheromone-guided policy optimization.

\noindent\textbf{Supervised warm-up.}
We first pre-train the policy with a next-tool prediction objective:
\begin{equation}
\mathcal{L}^{\text{SL}}=-\log \pi_\theta(a_t^*\mid s_t),
\label{eq:sl}
\end{equation}
where $a_t^*$ is the reference next tool. This stage ignores pheromone guidance and moves the policy away from random tool selection. By establishing tool-selection capabilities, the model can generate plausible initial tool-use trajectories that facilitate future success.

\noindent\textbf{Pheromone as guidance.}
Building on this initialization, we introduce a pheromone-guided exploration strategy to navigate long-horizon tool-ues task completion.
Specifically, we incorporate pheromone as an explicit transition prior during trajectory generation, increasing the likelihood of collecting informative tool-use trajectories for policy learning.
This pheromone-guided trajectory generation aligns well with group-based policy optimization \cite{group1,group2}, where learning signals depend on the quality of multiple rollouts per prompt. 
We adopt Group Relative Policy Optimization (GRPO) \cite{grpo} as the optimization backbone, since GRPO computes relative advantages within a rollout group. 
In long-horizon tool planning with a large tool space, unguided sampling can yield groups dominated by uniformly low-quality trajectories, weakening the relative learning signal. Pheromone-guided sampling increases the likelihood that each group contains higher-potential trajectories and meaningful contrasts, enabling GRPO to better leverage its group-based advantage to update the policy.
We now present the pheromone-guided sampling rule and the objective used to update the policy.

\noindent\textbf{Pheromone-guided sampling.}
At each step $t$, the policy network generates $\pi_\theta(a_t \mid s_t)$ given the current state $s_t$. To incorporate historical experience, we integrate the policy with pheromone as an explicit transition prior. 
Specifically, we first select the top-$K$ candidates from $\pi_\theta$ as the sampling support, and then sample the next tool according to the pheromone-guided policy:
\begin{equation}
p(a_t \mid s_t, a_{t-1}) \propto \exp \bigl( \log \pi_\theta(a_t \mid s_t) + \beta \log \tau_{\text{Tool}}(a_{t-1}, a_t \mid x) \bigr).
\label{eq:sampling}
\end{equation}
Once the base tool $a_t$ is selected, the agent selects an argument invocation
$u_t \in \mathcal{V}_{\text{Arg}}(a_t)$ according to $\tau_{\text{Arg}}(a_t, u_t \mid x)$.
This argument choice is not parameterized by $\pi_\theta$ and is not updated during policy optimization.
Here $(a_t, u_t)$ follows the tool-to-invocation association defined in Sec.~\ref{sec:tool-graph}.

\noindent\textbf{Policy optimization.}
For each training instance, we sample $M$ tool-use trajectories using the pheromone-guided distribution in Eq.~\eqref{eq:sampling}, obtaining returns $\{R_m\}_{m=1}^{M}$.
We compute group-relative advantages:
\begin{gather}
A_m = \frac{R_m - \bar{R}}{\sigma_R + \epsilon_{\mathrm{adv}}}, 
\qquad \text{where}\quad
\bar{R} = \frac{1}{M} \sum_{m=1}^M R_m,
\label{eq:adv}
\end{gather}
where $\epsilon_{\mathrm{adv}}$ is a small constant for numerical stability.
The policy $\pi_\theta$ is updated by minimizing the following objective:
\begin{gather}
\nonumber
\mathcal{L}^{\text{PG}} = -\mathbb{E} \left[
\min \left(
r_t(\theta) A_m,\,
\mathrm{clip}(r_t(\theta), 1-\epsilon_{\mathrm{clip}}, 1+\epsilon_{\mathrm{clip}}) A_m
\right)
\right],
\end{gather}
where $r_t(\theta) = \pi_\theta(a_t \mid s_t) / \pi_{\theta_{\text{old}}}(a_t \mid s_t)$ denotes the probability ratio at step $t$, and $\epsilon_{\mathrm{clip}}$ is the clipping threshold.

\noindent\textbf{Progressive pheromone updates.}
Having introduced pheromone-guided sampling and the policy optimization objective, we apply them  through a progressive training schedule that blends supervision and exploration to accumulate pheromone as an explicit transition prior.
We introduce a learning curriculum over horizon lengths, starting from short horizons and gradually increasing to longer ones, and employ an oracle probability $p_{\text{tf}}$ decaying over time. Specifically, at each step $t$, the action $a_t$ is determined as:
\begin{equation}
a_t =
\begin{cases}
a_t^*, & \text{with prob. } p_{\text{tf}}, \\
\text{sampled from Eq.~\eqref{eq:sampling}}, & \text{with prob. } 1-p_{\text{tf}},
\end{cases}
\label{eq:tf}
\end{equation}
where $a_t^*$ is the reference action.
This keeps rollouts close to reference trajectories in the early phase, increasing the chance of collecting verified successful tool-use trajectories for pheromone updates in longer horizons.
We employ a mixed objective that shifts from imitation learning to policy optimization:
\begin{equation}
\mathcal{L}^{\text{mixed}} = \lambda \mathcal{L}^{\text{SL}} + (1-\lambda) \mathcal{L}^{\text{PG}} - \gamma \mathcal{H}[\pi_\theta],
\label{eq:mixed}
\end{equation}
where $\lambda$ is annealed downwards and $\mathcal{H}[\pi_{\theta}]$ denotes the entropy of $\pi_{\theta}$.
Accordingly, the explicit transition prior (pheromone) is updated from partially autonomous verified successful trajectories, making it increasingly reflect policy-selected transitions, instead of being dominated by purely teacher-forced ones.
We concurrently increase the pheromone-controlling hyperparameter $\beta$ in Eq.~\eqref{eq:sampling} and the task-dependence weight $w$ in Eq.~\eqref{eq:fuse-pher} as the number of accumulated trajectories grows, turning the initial pheromone values into a reliable transition signal.

\noindent\textbf{Full-pheromone rollouts.}
As $p_{\text{tf}}$ decays and $\lambda$ is annealed downwards, training transitions to fully autonomous rollouts with full pheromone influence:
\begin{equation}
\tilde{\pi}(a \mid s_t) \propto \exp \bigl( \log \pi_\theta(a \mid s_t) + \beta_{\text{max}} \log \tau_{a_{t-1}, a} \bigr),
\label{eq:stage3_sampling}
\end{equation}
where $\beta_{\text{max}}$ is the maximum pheromone influence scheduled to be reached, and $\tau_{a_{t-1},a}$ is a shorthand for the fused tool-transition pheromone $\tau_{\text{Tool}}(a_{t-1}, a_t \mid x)$. Similarly, argument-invocation $u_t$ is selected using the same pheromone-guided rule with $\tau_\text{Arg}(a_t, u_t \mid x)$.
The updated pheromone serves as an explicit transition prior during trajectory generation, encouraging the policy to reproduce historically successful transitions and argument invocations. 
Policy optimization is performed on tool-use trajectories collected under pheromone guidance. In parallel, pheromone updates follow the deposition and evaporation rule defined in Eq.~\eqref{eq:agn}, where evaporation reduces the influence of outdated statistics over time.

Collectively, training begins with supervised warm-up that provides a stable initialization for next-tool selection. It then proceeds with pheromone-guided reinforcement learning that uses pheromone as an explicit transition prior during trajectory generation and optimizes the policy with GRPO on tool-use trajectories collected under pheromone guidance. The training schedule gradually reduces oracle guidance via the decayed $p_{\text{tf}}$ and anneals $\lambda$ downward, and the horizon curriculum increases task difficulty from short to long. This progression supports progressive pheromone accumulation from partially guided verified successes and later enables fully pheromone-guided trajectory generation for complex long-horizon tool-use trajectories.

\section{Experiments}

\subsection{Experiment Setup}
\label{sec:exp-setup}

\noindent\textbf{Dataset.}
We conduct experiments on three long-horizon tool-use benchmarks: \textbf{Toolathlon}~\cite{toolathlon}, a benchmark built on MCP servers with hundreds of tools across diverse domains; \textbf{TRAJECT-Bench}~\cite{traject}, grounded in executable production-style APIs; and \textbf{TOUCAN}~\cite{toucan}, a large-scale tool-agent dataset synthesized from MCP environments.
Each instance is converted into an episode containing a task description, a reference tool-use trajectory, and the corresponding execution outputs.
We filter episodes with verified successful trajectories and map each recorded tool call to our action representation by separating the base tool from its argument invocation.
We construct fixed-seed training, validation, and test splits and report results on the test set.

\noindent\textbf{Comparing Methods.}
We compare with a diverse set of representative tool-use baselines. 
\textbf{Standard Prompting and Planning} methods include the reasoning-action interleaving ReAct~\cite{react} and global-oriented Plan-and-Solve~\cite{Plan-and-Solve}, alongside the Beyond-ReAct~\cite{beyond} planner-executor architecture. 
\textbf{Retrieval-Augmented Selection} baselines comprise Gorilla~\cite{gorilla}, ToolReAGt~\cite{toolreagt}, and the GCN-based Tool Graph Retriever~\cite{tgr}.
\textbf{RL-based Optimization} baselines consist of search-driven TreeRL~\cite{treerl}, outcome-oriented ToRL~\cite{torl}, and process-level StepTool~\cite{steptool}.
Finally, \textbf{Graph-based Frameworks} include GAP~\cite{gap} as well as GTool~\cite{gtool}, NaviAgent~\cite{naviagent}, and ToolNet~\cite{toolnet}, which leverage explicit tool-dependency structures for navigation or representation learning.
All experiments are conducted using \textbf{Qwen2.5-7B} \cite{qwen2.5} and \textbf{Llama3.1-8B} \cite{llama3.1} as backbones. All baselines operate in the same tool-and-invocation decision space and share the same execution interface.

\noindent\textbf{Evaluation Metrics.}
We evaluate performance using two metrics:
(1) \textbf{Match Ratio ($M \in [0,1]$)}, a sequence-level metric that measures the similarity between generated tool-use trajectories and reference trajectories in tool identity and ordering; and
(2) \textbf{Task Success Rate ($\mathrm{TSR} \in [0,1]$)}, an execution-level metric that measures whether the generated tool-use trajectory successfully completes the target task.
We evaluate TSR using GPT-4o as an outcome judge under a fixed evaluation protocol, where the judge determines whether the final tool-use outcome satisfies the task requirement.
Unlike Match Ratio, TSR evaluates the final task outcome and does not require the generated trajectory to exactly follow the reference ordering.
We report both metrics as percentages in the experiments.

\noindent\textbf{Implementation Details.}
We fine-tune the backbones using LoRA~\citep{lora} ($r=64$, $\alpha=128$) to adapt the policy for long-horizon tool planning.
Each state consists of the task description and a fixed-length truncated history of recent tool interactions.
For decoding, we employ temperature $0.7$, top-$K$ sampling with $K=20$, and random exploration probability $p_{\mathrm{rand}}=0.05$.
The pheromone influence $\beta$ is linearly scheduled to a maximum value $\beta_{\max}=0.7$.
We set the maximum trajectory length to $20$ and sample $M=5$ rollouts for optimization.
Complete experimental details are provided in Appendix~\ref{app:impl}.

\subsection{Experimental Results}
\label{sec:main-results}

\definecolor{highlightblue}{rgb}{0.92, 0.96, 1.0} 
\definecolor{categorygray}{gray}{0.92} 

\begin{table*}[t]
\centering
\scriptsize
\caption{\textbf{Main Results on Long-Horizon Tool-Use Benchmarks.} We report Match Ratio (Match R., \%) and Task Success Rate (TSR, \%) on Toolathlon, TRAJECT-Bench, and TOUCAN across two backbone models. Higher values ($\uparrow$) indicate better performance.}
\label{tab:main_results}
\setlength{\tabcolsep}{3.8pt}
\begin{tabular}{@{}lcccccccccccc@{}}
\toprule
\multicolumn{1}{c}{\multirow{3}{*}{\textbf{Method}}}
& \multicolumn{4}{c}{\textbf{Toolathlon}} 
& \multicolumn{4}{c}{\textbf{TRAJECT-Bench}} 
& \multicolumn{4}{c}{\textbf{TOUCAN}} \\
\cmidrule(lr){2-5} \cmidrule(lr){6-9} \cmidrule(lr){10-13}
& \multicolumn{2}{c}{\textbf{Qwen2.5-7B}} & \multicolumn{2}{c}{\textbf{Llama3.1-8B}}
& \multicolumn{2}{c}{\textbf{Qwen2.5-7B}} & \multicolumn{2}{c}{\textbf{Llama3.1-8B}}
& \multicolumn{2}{c}{\textbf{Qwen2.5-7B}} & \multicolumn{2}{c}{\textbf{Llama3.1-8B}} \\
\cmidrule(lr){2-3} \cmidrule(lr){4-5}
\cmidrule(lr){6-7} \cmidrule(lr){8-9}
\cmidrule(lr){10-11} \cmidrule(lr){12-13}
& Match R. & TSR & Match R. & TSR
& Match R. & TSR & Match R. & TSR
& Match R. & TSR & Match R. & TSR \\
\midrule

\rowcolor{categorygray}
\multicolumn{13}{@{}l}{\textit{Standard Prompting \& Planning Strategies}} \\
ReAct & 5.31 & 2.92 & 5.15 & 2.48 & 15.32 & 9.20 & 9.16 & 7.01 & 11.64 & 8.92 & 7.03 & 5.88 \\
Plan-and-Solve & 6.42 & 4.61 & 4.27 & 3.96 & 12.44 & 11.35 & 9.05 & 5.54 & 12.71 & 10.15 & 7.42 & 5.12 \\
Beyond ReAct & 3.85 & 3.10 & 5.23 & 4.18 & 11.33 & 10.74 & 15.61 & 14.20 & 12.92 & 10.84 & 7.35 & 6.54 \\
\midrule

\rowcolor{categorygray}
\multicolumn{13}{@{}l}{\textit{Retrieval-Augmented Generation (RAG)}} \\
Gorilla & 6.60 & 3.26 & 5.78 & 3.65 & 29.04 & 22.62 & 19.38 & 11.42 & 22.32 & 16.05 & 18.99 & 16.94 \\
ToolReAGt & 4.05 & 4.42 & 3.45 & 3.17 & 26.69 & 20.11 & 26.84 & 18.67 & 23.49 & 18.77 & 15.79 & 10.76 \\
Tool Graph Retriever & 5.09 & 3.74 & 4.80 & 3.58 & 12.72 & 9.40 & 11.54 & 9.95 & 1.50 & 2.12 & 1.50 & 1.83 \\
\midrule

\rowcolor{categorygray}
\multicolumn{13}{@{}l}{\textit{Reinforcement Learning}} \\
ToRL & 10.23 & 8.65 & 9.41 & 3.46 & 28.15 & 16.05 & 20.18 & 12.36 & 19.32 & 18.90 & 14.35 & 10.94 \\
StepTool & 14.33 & 12.08 & 11.58 & 7.87 & 35.22 & 23.18 & 26.31 & 20.12 & 19.75 & 20.86 & 14.46 & 15.30 \\
TreeRL & 21.26 & 18.91 & 18.78 & 11.42 & 35.45 & 27.82 & 26.33 & 20.04 & 30.65 & 18.74 & 20.20 & 18.62 \\
\midrule

\rowcolor{categorygray}
\multicolumn{13}{@{}l}{\textit{Graph-based Navigation Models}} \\
ToolNet & 9.01 & 4.83 & 7.30 & 4.15 & 20.65 & 11.84 & 18.52 & 13.87 & 16.15 & 12.10 & 13.53 & 10.15 \\
GAP & 13.50 & 10.76 & 12.17 & 9.54 & 27.60 & 21.10 & 21.01 & 13.45 & 23.80 & 17.86 & 15.43 & 13.62 \\
NaviAgent & 15.28 & 13.94 & 11.60 & 10.86 & 25.33 & 20.78 & 19.40 & 14.83 & 23.10 & 20.44 & 16.20 & 16.05 \\
GTool & 20.60 & 17.82 & 16.50 & 14.22 & 32.50 & 28.06 & 26.20 & 21.78 & 27.56 & 19.35 & 19.50 & 13.33 \\
\midrule

\rowcolor{highlightblue}
\textbf{PhGPO (Ours)} & \textbf{25.25} & \textbf{20.34} & \textbf{22.83} & \textbf{19.76} & \textbf{41.62} & \textbf{32.18} & \textbf{31.84} & \textbf{22.66} & \textbf{35.17} & \textbf{26.28} & \textbf{24.67} & \textbf{19.86} \\
\bottomrule
\end{tabular}
\end{table*}

\noindent\textbf{Overall performance compared with prior methods.}
Table~\ref{tab:main_results} shows that \textbf{PhGPO achieves the best performance} on Toolathlon, TRAJECT-Bench, and TOUCAN under both Qwen2.5-7B and Llama3.1-8B.
Compared with the strongest baselines, PhGPO consistently improves both Match Ratio and Task Success Rate, indicating that its gains appear in both reference-trajectory alignment and final task completion.
The improvements are observed across different benchmarks and backbones, suggesting that the method is not tied to a specific task distribution or base model.

\noindent\textbf{Compared with prompting and retrieval-augmented baselines.}
Prompting and planning baselines remain substantially below learning-based methods, reflecting the difficulty of maintaining coherent multi-step tool transitions through in-context reasoning alone.
Retrieval-augmented methods can narrow the candidate tool space and improve local grounding, but their Task Success Rate remains limited on long-horizon episodes.
This suggests that locally relevant tool retrieval does not necessarily lead to globally consistent tool-use trajectories, where later decisions must remain compatible with earlier tools, intermediate states, and argument invocations.

\noindent\textbf{Compared with graph-based and RL-based optimization baselines.}
Graph-based navigation models improve over prompting baselines by using explicit tool structure, while RL-based methods further benefit from learning with execution outcomes.
PhGPO outperforms these methods by using pheromone as an explicit transition prior during rollout generation and policy optimization.
This prior biases sampling toward tool-transition edges and argument-invocation patterns that recur in verified successful trajectories, which improves multi-step consistency and reduces cascading errors.
The higher Task Success Rate further shows that the improvement is not limited to reference-path imitation, but also leads to more reliable task completion.

\begin{figure*}[ht]
  \begin{center}
    \centerline{\includegraphics[width=0.85\textwidth]{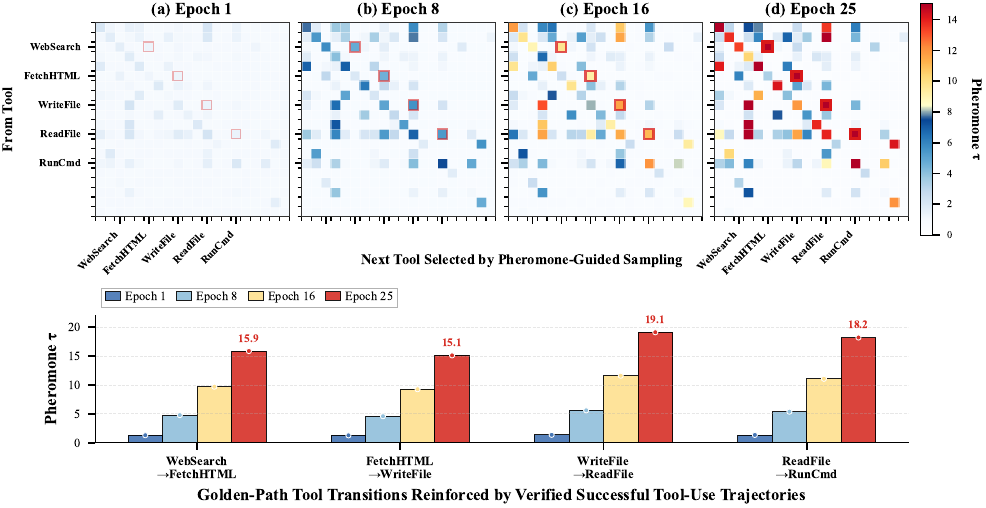}}
    \caption{\textbf{Emergence of the reference chain in pheromone transitions.}
    As training proceeds, the transition matrix becomes increasingly concentrated on reference edges (red), while unrelated transitions fade due to evaporation, and pheromone values on the reference chain increase steadily.}
    \label{pheromone}
  \end{center}
  \vskip -0.6cm
\end{figure*}

\subsection{Ablation Studies}
\label{sec:ablation}
We conduct ablation studies on Qwen2.5-7B using the same Toolathlon split, training budget, and evaluation protocol as the main experiment.
All variants share the same backbone and optimization settings, with only one component modified at a time.
We compare PhGPO with five variants:
\textbf{(1) w/o Pheromone}, which sets $\beta{=}0$ throughout training;
\textbf{(2) w/o Mixed Curriculum}, which removes the oracle-guided curriculum and accumulates pheromones only from fully policy-driven rollouts;
\textbf{(3) Static Prior}, which freezes pheromones after the progressive training pipeline;
\textbf{(4) w/o Evaporation}, which sets $\rho{=}0$ so pheromones only accumulate;
and \textbf{(5) w/o Task-dependent Pheromones}, which uses only task-agnostic pheromones.

\definecolor{highlightblue}{rgb}{0.92, 0.96, 1.0}

\begin{table*}[t]
\centering
\scriptsize
\caption{\textbf{Ablation studies on Toolathlon.} We report Match Ratio (Match R., \%) and Task Success Rate (TSR, \%) on the test set. Drop columns are computed relative to PhGPO (Full), where larger drops indicate stronger contribution of the removed component.}
\label{tab:ablation_qwen7b}
\setlength{\tabcolsep}{5.5pt}
\begin{tabular}{@{}llcccc@{}}
\toprule
\textbf{Variant} & \textbf{Variant role} & \textbf{Match R.$\uparrow$} & \textbf{Drop M.$\downarrow$} & \textbf{TSR$\uparrow$} & \textbf{Drop TSR$\downarrow$} \\
\midrule
w/o Pheromone ($\beta{=}0$ throughout) 
& removes explicit transition prior 
& 21.23 & 4.02 & 16.72 & 3.62 \\

w/o Mixed Curriculum 
& removes progressive oracle guidance 
& 19.34 & 5.91 & 14.86 & 5.48 \\

Static Prior 
& freezes pheromone updates 
& 23.93 & 1.32 & 18.71 & 1.63 \\

w/o Evaporation ($\rho{=}0$) 
& removes decay of stale statistics 
& 21.72 & 3.53 & 16.28 & 4.06 \\

w/o Task-dependent Pheromones 
& removes task-matched edge memory 
& 22.45 & 2.80 & 17.35 & 2.99 \\
\midrule
\rowcolor{highlightblue}
\textbf{PhGPO (Full)} 
& \textbf{all components enabled} 
& \textbf{25.25} & -- & \textbf{20.34} & -- \\
\bottomrule
\end{tabular}
\end{table*}

Table~\ref{tab:ablation_qwen7b} shows that every component contributes to both reference-trajectory alignment and final task completion.
Removing the Mixed Curriculum causes the largest drop, indicating that early guided rollouts help accumulate reliable pheromone statistics before training becomes fully policy driven.
Disabling pheromone guidance also substantially degrades performance, showing that the explicit transition prior is important for long-horizon trajectory generation.
Freezing pheromones or removing evaporation underperforms the full model, which supports the need for online updates and decay to keep the prior aligned with the evolving policy.
Finally, removing task-dependent pheromones reduces both Match Ratio and TSR, suggesting that task-matched transition evidence helps improve tool-use decisions for the current instance.

\subsection{Visualization of transition structure}
To qualitatively examine the transition structure learned by PhGPO, we present a representative long-horizon task involving sequential web retrieval and file manipulation. Figure~\ref{pheromone} visualizes the evolution of the pheromone transition matrix during training.

At the beginning of training, the matrix is diffuse and no clear transition pattern is emphasized. As training proceeds, pheromone gradually concentrates on a small subset of transitions. In particular, the reference chain (\textit{WebSearch} $\to$ \textit{FetchHTML} $\to$ \textit{WriteFile} $\to$ \textit{ReadFile} $\to$ \textit{RunCmd}) is repeatedly reinforced, while unsupported transitions fade under evaporation. By Epoch 25, the reference transitions are clearly highlighted, showing that PhGPO progressively accumulates and reuses validated transition patterns.

\section{Conclusion and Limitations}
\label{sec:conclusion}

We proposed PhGPO, a pheromone-guided policy optimization framework for long-horizon tool planning. PhGPO maintains a pheromone-based explicit transition prior over tool-transition edges and argument invocations, updates it through deposition and evaporation, and incorporates it into rollout sampling to reuse historically successful tool-use patterns. Experiments show consistent improvements in both Match Ratio and Task Success Rate, indicating benefits for reference-trajectory alignment and final task completion. We hope this work encourages future research on explicit experience reuse for tool-use agents. Future extensions may study broader verification settings, richer argument-invocation modeling, and more dynamic tool environments with evolving APIs.

\bibliographystyle{unsrtnat}
\bibliography{paper}

\newpage
\appendix

\section{Theoretical Motivation of Pheromone-guided Sampling}
\label{app:theory}

This appendix provides a principled view of the pheromone-guided sampling distribution in Eq.~\eqref{eq:sampling}.
The key idea is that the sampling rule can be derived as the closed-form solution of an entropy-regularized consensus problem that combines the neural policy with an \emph{explicit transition prior} induced by pheromone.

\subsection{Setup}
At decision step $t$, the policy produces a distribution $\pi_\theta(a\mid s_t)$ over tools.
Following Sec.~\ref{sec:training}, we restrict sampling to a finite support $\mathcal{A}_t$ constructed from the Top-$K$ candidates under $\pi_\theta(\cdot\mid s_t)$.
We use pheromone as an explicit transition prior through the fused tool-transition pheromone $\tau_{\text{Tool}}(a_{t-1},a\mid x)$ defined in Sec.~\ref{sec:pheromone}, which already incorporates task-agnostic and task-dependent evidence via Eq.~\eqref{eq:fuse-pher}.
To ensure $\log \tau_{\text{Tool}}$ is well-defined, we clip pheromone values:
\begin{equation}
\bar{\tau}(a_{t-1},a\mid x)
=
\mathrm{clip}\bigl(\tau_{\text{Tool}}(a_{t-1},a\mid x),[\tau_{\min},\tau_{\max}]\bigr),
\qquad \tau_{\min}>0.
\label{eq:tau_clip_app}
\end{equation}
We define the pheromone-induced explicit transition prior over the candidate set as
\begin{equation}
p_{\tau}(a \mid a_{t-1},x)
\propto
\bigl[\bar{\tau}(a_{t-1},a\mid x)\bigr]^{\beta},
\qquad a\in\mathcal{A}_t,
\label{eq:prior_app}
\end{equation}
where $\beta\ge 0$ controls the strength of the explicit transition prior.
The pheromone-guided sampling distribution used in our rollouts is
\begin{equation}
\tilde{\pi}(a\mid s_t,a_{t-1},x)
\propto
\exp\!\Big(\log \pi_\theta(a\mid s_t) + \beta \log \bar{\tau}(a_{t-1},a\mid x)\Big),
\qquad a\in\mathcal{A}_t,
\label{eq:fuse_app}
\end{equation}
which is equivalent to the multiplicative form $\tilde{\pi}\propto \pi_\theta(\cdot\mid s_t)\,p_\tau(\cdot\mid a_{t-1},x)$ on the support $\mathcal{A}_t$.

\subsection{Entropy-regularized consensus objective}
We derive $\tilde{\pi}$ by solving for a distribution $q$ over the candidate set $\mathcal{A}_t$ that agrees with both the neural policy and the explicit transition prior, while retaining entropy:
\begin{equation}
\max_{q}\;
\mathbb{E}_{a\sim q}\!\Big[\log \pi_\theta(a\mid s_t)+ \beta \log \bar{\tau}(a_{t-1},a\mid x)\Big]
+\eta\,\mathcal{H}(q),
\quad
\text{s.t. } \sum_{a\in\mathcal{A}_t} q(a)=1,\; q(a)\ge 0,
\label{eq:obj_app}
\end{equation}
where $\mathcal{H}(q)=-\sum_{a\in\mathcal{A}_t} q(a)\log q(a)$ is Shannon entropy and $\eta>0$ is an entropy weight.

\subsection{Closed-form solution via Lagrange multipliers}
Introduce a Lagrange multiplier $\gamma$ for the normalization constraint and write the Lagrangian
\begin{align}
\mathcal{L}(q,\gamma)
&= \sum_{a\in\mathcal{A}_t} q(a)\Big(\log \pi_\theta(a\mid s_t)+ \beta \log \bar{\tau}(a_{t-1},a\mid x)\Big) \nonumber \\
&\quad +\eta\Big(-\sum_{a\in\mathcal{A}_t} q(a)\log q(a)\Big) \nonumber \\
&\quad +\gamma\Big(\sum_{a\in\mathcal{A}_t} q(a)-1\Big).
\label{eq:lagrangian_app}
\end{align}
Taking partial derivatives with respect to $q(a)$ and setting them to zero gives
\begin{equation}
\log \pi_\theta(a\mid s_t)+ \beta \log \bar{\tau}(a_{t-1},a\mid x)
-\eta\bigl(1+\log q(a)\bigr)
+\gamma
=0.
\label{eq:stationary_app}
\end{equation}
Rearranging yields
\begin{equation}
\log q(a)
=
\frac{1}{\eta}\Big(\log \pi_\theta(a\mid s_t)+ \beta \log \bar{\tau}(a_{t-1},a\mid x)\Big)
+\frac{\gamma}{\eta}-1,
\label{eq:logq_app}
\end{equation}
which implies the optimal solution
\begin{equation}
q^*(a)
\propto
\exp\!\Big(\tfrac{1}{\eta}\big[\log \pi_\theta(a\mid s_t)+ \beta \log \bar{\tau}(a_{t-1},a\mid x)\big]\Big),
\qquad a\in\mathcal{A}_t.
\label{eq:sol_app}
\end{equation}
Setting $\eta=1$ recovers Eq.~\eqref{eq:fuse_app}, which matches our implementation in Eq.~\eqref{eq:sampling}.

\subsection{Equivalent KL projection view}
The same objective admits an information-geometric interpretation.
Let
\begin{equation}
r(a)\propto \exp\!\Big(\log \pi_\theta(a\mid s_t)+ \beta \log \bar{\tau}(a_{t-1},a\mid x)\Big),
\qquad a\in\mathcal{A}_t.
\label{eq:r_app}
\end{equation}
Expanding $\mathrm{KL}(q\|r)=\sum_{a\in\mathcal{A}_t} q(a)\log\frac{q(a)}{r(a)}$ gives
\begin{equation}
-\mathrm{KL}(q\|r)
=
\mathbb{E}_{a\sim q}\big[\log \pi_\theta(a\mid s_t)+ \beta \log \bar{\tau}(a_{t-1},a\mid x)\big]
+\mathcal{H}(q)
+\text{const}.
\label{eq:kl_expand_app}
\end{equation}
For $\eta=1$, maximizing Eq.~\eqref{eq:obj_app} is equivalent, up to constants, to the KL projection
\begin{equation}
\min_q \ \mathrm{KL}\!\left(q \,\middle\|\, \tfrac{1}{Z}\exp\!\big(\log \pi_\theta(\cdot\mid s_t)+ \beta \log \bar{\tau}(a_{t-1},\cdot\mid x)\big)\right),
\label{eq:kl_app}
\end{equation}
where $Z$ is the normalizer over $\mathcal{A}_t$.
This view clarifies the role of pheromone as an explicit transition prior in log space: actions must be supported by the neural policy and by the accumulated transition statistics encoded in $\bar{\tau}$, which increases the probability of historically successful tool-transition edges under the current task context.

\begin{algorithm}
  \caption{PhGPO Training Loop with Task-dependent Pheromone Fusion}
  \label{alg:PhGPO}
  \begin{algorithmic}[1]
    \STATE {\bfseries Input:}
    dataset $\mathcal{D}$ with task $x$ and reference trajectory $\{a_t^*\}$;
    horizon curriculum $\{L_s\}_{s=1}^{S}$;
    schedules $p_{\text{tf}}(s), \lambda(s), \beta(s), w(s)$;
    group size $M$;
    tool graph $\mathcal{G}{=}(\mathcal{V}_{\text{Tool}},\mathcal{E})$ and invocation edges $\mathcal{E}_{\text{arg}}$;
    sentence encoder $\phi(\cdot)$;
    ACO parameters $(\rho,\alpha,\tau_{\min},\tau_{\max},\tau_0)$;
    $\theta_{\text{sim}}, n_{\min}, \epsilon_{\mathrm{dep}}$.
    \STATE {\bfseries Output:}
    policy $\pi_\theta$; pheromones $\tau^{\text{agn}}_{\text{Tool}}, \tau^{\text{agn}}_{\text{Arg}}$; memory banks $\{\mathcal{M}\}$.

    \STATE Initialize $\pi_\theta$; set $\tau^{\text{agn}}_{\text{Tool}}, \tau^{\text{agn}}_{\text{Arg}}\leftarrow\tau_0$; initialize all memory banks as empty.

    \STATE {\footnotesize \textit{Subroutine} $\mathrm{ComputeFusedPheromone}(\tau^{\text{agn}}, \mathcal{M}, \bm{e}_x, w)$:}
    \STATE {\footnotesize Retrieve $\mathcal{M}(x)=\{(\bm{e},q_{\tau})\in\mathcal{M}\mid \mathrm{sim}(\bm{e}_x,\bm{e})\ge \theta_{\text{sim}}\}$.}
    \STATE {\footnotesize Compute $\tau^{\text{dep}}(\cdot\mid x)$ by Eq.~\eqref{eq:dep} and confidence $c(x)$ by Eq.~\eqref{eq:confidence}.}
    \STATE {\footnotesize Return fused $\tau(\cdot\mid x)$ by Eq.~\eqref{eq:fuse-pher}, clipped to $[\tau_{\min},\tau_{\max}]$.}
    \FOR{epochs}
      \STATE Update $\pi_\theta$ by minimizing $\mathcal{L}^{\text{SL}}$ in Eq.~\eqref{eq:sl}.
    \ENDFOR

    \FOR{$s=1$ {\bfseries to} $S$}
      \STATE $L\leftarrow L_s;\ p_{\text{tf}}\leftarrow p_{\text{tf}}(s);\ \lambda\leftarrow\lambda(s);\ \beta\leftarrow\beta(s);\ w\leftarrow w(s)$.
      \FOR{each $(x,\{a_t^*\})\in\mathcal{D}$}
        \STATE $\bm{e}_x \leftarrow \phi(x)$.
        \FOR{$m=1$ {\bfseries to} $M$}
          \STATE $a_0\leftarrow\langle\texttt{START}\rangle$; initialize $\xi_m$.
          \FOR{$t=1$ {\bfseries to} $L$}
            \STATE Build state $s_t$ from $(x,\xi_m)$.
            \IF{$\mathrm{rand()}<p_{\text{tf}}$}
              \STATE $a_t\leftarrow a_t^*$ \COMMENT{Eq.~\eqref{eq:tf}}
            \ELSE
              \STATE Form candidate set $\mathcal{A}_t$ as Top-$K$ under $\pi_\theta(\cdot\mid s_t)$.
              \FOR{each $a\in\mathcal{A}_t$}
                \STATE $\tau_{\text{Tool}}(a_{t-1},a\mid x)\leftarrow
                \mathrm{ComputeFusedPheromone}\bigl(\tau^{\text{agn}}_{\text{Tool}}(a_{t-1},a),\mathcal{M}_{a_{t-1},a},\bm{e}_x,w\bigr)$.
              \ENDFOR
              \STATE Sample $a_t$ using Eq.~\eqref{eq:sampling} with $\beta$ on support $\mathcal{A}_t$.
            \ENDIF
            \STATE Let $\mathcal{U}_t\leftarrow \mathcal{V}_{\text{Arg}}(a_t)$.
            \FOR{each $u\in\mathcal{U}_t$}
              \STATE $\tau_{\text{Arg}}(a_t,u\mid x)\leftarrow
              \mathrm{ComputeFusedPheromone}\bigl(\tau^{\text{agn}}_{\text{Arg}}(a_t,u),\mathcal{M}_{a_t,u},\bm{e}_x,w\bigr)$.
            \ENDFOR
            \STATE Sample $u_t$ by the analogous pheromone-guided rule; execute $(a_t,u_t)$ and append to $\xi_m$.
          \ENDFOR
          \STATE Compute return $R_m$ and pheromone-deposition score $q_{\tau}(\xi_m)\in[0,1]$ (match ratio).
        \ENDFOR

        \STATE Compute advantages by Eq.~\eqref{eq:adv}.
        \STATE Update $\pi_\theta$ using the mixed objective in Eq.~\eqref{eq:mixed}.

        \FOR{each rollout $\xi_m$ that is verified successful}
          \FOR{each traversed tool-transition edge $(v_i,v_j)$ in $\xi_m$}
            \STATE Update $\tau^{\text{agn}}_{\text{Tool}}(v_i,v_j)$ by Eq.~\eqref{eq:agn}.
            \STATE Store $(\bm{e}_x,q_{\tau}(\xi_m))$ into $\mathcal{M}_{ij}$.
          \ENDFOR
          \FOR{each traversed invocation edge $(v_i,u)$ in $\xi_m$}
            \STATE Update $\tau^{\text{agn}}_{\text{Arg}}(v_i,u)$ by Eq.~\eqref{eq:agn}.
            \STATE Store $(\bm{e}_x,q_{\tau}(\xi_m))$ into $\mathcal{M}_{i,u}$.
          \ENDFOR
        \ENDFOR
      \ENDFOR
    \ENDFOR
  \end{algorithmic}
\end{algorithm}

\section{Algorithm Implementation Details and Training Mechanisms}
\label{app:impl}

\subsection{Base model adaptation}
\label{app:base}

We instantiate PhGPO on instruction-tuned backbones used in our main experiments, including Qwen2.5-7B and Llama3.1-8B.
We apply LoRA for parameter-efficient adaptation by freezing backbone weights and inserting low-rank adapters with rank $r{=}64$ and scaling $\alpha{=}128$ into standard projection modules.
All stages use the tokenizer and chat template of the corresponding backbone, and tool planning is trained as a generative decision process over the same two-layer action space described in Sec.~\ref{sec:tool-graph}.
Our code is available at \url{https://anonymous.4open.science/r/PhGPO-Pheromone-Guided-Policy-Optimization-for-Long-Horizon-Tool-Planning-CB1C}.

\subsection{Progressive training pipeline}
\label{app:pipeline}

Training follows the progressive pipeline described in Sec.~\ref{sec:training}.
It begins with supervised warm-up using the next-tool prediction objective in Eq.~\eqref{eq:sl}, which provides a stable initialization for tool selection.
After warm-up, training uses a progressive schedule that mixes oracle guidance with pheromone-guided sampling.
At each decision step, the agent takes the reference action with probability $p_{\text{tf}}$; otherwise it samples actions from the pheromone-guided distribution in Eq.~\eqref{eq:sampling}, as specified in Eq.~\eqref{eq:tf}.
Policy updates use the mixed objective in Eq.~\eqref{eq:mixed}, where $\lambda$ is annealed downward to shift learning from imitation toward policy optimization.
In parallel, a horizon curriculum increases the rollout horizon from shorter to longer trajectories.
We also schedule $\beta$ in Eq.~\eqref{eq:sampling} and $w$ in Eq.~\eqref{eq:fuse-pher} upward as more verified trajectories are accumulated, which increases the influence of the explicit transition prior and task-dependent evidence.
When $p_{\text{tf}}$ approaches zero and $\lambda$ becomes small, rollouts become fully policy-driven under pheromone guidance, and optimization follows the policy objective described in Sec.~\ref{sec:training}.
For group-based learning, we use $M{=}5$ rollouts per instance, consistent with Sec.~\ref{sec:exp-setup}.

\subsection{Mixed training objective under teacher forcing}
\label{app:mixed-objective}

During progressive training pipeline, rollouts are generated under a mixed behavior rule that combines oracle actions and pheromone-guided sampling.
Let $a_t^\ast$ denote the reference action provided by the benchmark instance, and let $\tilde{\pi}$ denote the pheromone-guided sampling distribution in Eq.~\eqref{eq:sampling}.
The behavior distribution used for rollout generation is
\begin{equation}
\pi_{\text{mix}}(a_t\mid s_t)
=
p_{\text{tf}}\cdot \delta(a_t = a_t^\ast)
\;+\;
(1-p_{\text{tf}})\cdot \tilde{\pi}(a_t\mid s_t),
\label{eq:mix-behavior}
\end{equation}
where $\delta(\cdot)$ is a point mass on the oracle action.
We record the executed action $a_t$ at every step, together with its log-probability under the current policy at rollout time, $\log \pi_{\theta_{\text{old}}}(a_t\mid s_t)$.

Policy optimization uses GRPO with a PPO-style clipped surrogate.
For each training instance, we sample a group of $M$ tool-use trajectories and compute their returns $\{R_m\}_{m=1}^M$ (Sec.~\ref{app:reward-structure}).
We normalize returns within the group to obtain advantages $A_m$ as in Eq.~\eqref{eq:adv}.
The policy-gradient loss is computed on executed actions under $\pi_{\text{mix}}$:
\begin{equation}
\begin{gathered}
\mathcal{L}_{\text{PG}}
= -\mathbb{E}_{m,t}\!\left[
\min\!\Bigl(r_{m,t}(\theta)A_m,\;
\mathrm{clip}(r_{m,t}(\theta),1-\epsilon_{\mathrm{clip}},1+\epsilon_{\mathrm{clip}})A_m\Bigr)
\right], \\
r_{m,t}(\theta)=\frac{\pi_{\theta}(a_{m,t}\mid s_{m,t})}{\pi_{\theta_{\text{old}}}(a_{m,t}\mid s_{m,t})}.
\end{gathered}
\label{eq:pg-tf}
\end{equation}
When an action is oracle-selected, Eq.~\eqref{eq:pg-tf} still evaluates its probability under $\pi_\theta$, since that action is the one executed in the environment under $\pi_{\text{mix}}$.
In parallel, the supervised term in Eq.~\eqref{eq:mixed} is computed as a next-action cross-entropy against $a_t^\ast$.
As $p_{\text{tf}}$ is annealed toward zero, the rollout distribution approaches fully policy-driven pheromone-guided sampling, and updates become dominated by autonomous decisions.

\subsection{Pheromone and memory}
\label{app:pheromone-bookkeeping}

PhGPO maintains pheromone on two types of edges: tool-transition edges $(v_i, v_j)\in\mathcal{E}$ and tool-to-invocation edges $(v_i, v_i^{(k)})\in\mathcal{E}_{\text{arg}}$ defined in Sec.~\ref{sec:tool-graph}.
We maintain task-agnostic pheromones $\tau^{\text{agn}}_{\text{Tool}}$ and $\tau^{\text{agn}}_{\text{Arg}}$, and we clip pheromone values to $[\tau_{\min},\tau_{\max}]$ as described in Sec.~\ref{sec:pheromone}.

\paragraph{Task embeddings and sentence encoder.}
Task-dependent pheromones use a fixed sentence encoder $\phi$ to embed the task text $x$ into a dense vector $\bm{e}=\phi(x)$.
In our reference implementation, $\phi$ is instantiated with \texttt{SentenceTransformer(all-MiniLM-L6-v2)}.
Edge memory banks store tuples $(\bm{e},q_{\tau})$ for each edge, where $q_{\tau}$ is the pheromone-deposition score of the corresponding tool-use trajectory (Sec.~\ref{app:reward-structure}).

We additionally maintain edge-indexed memory banks for task-dependent pheromones.
For a tool-transition edge $(v_i,v_j)$, its memory bank $\mathcal{M}_{ij}$ stores tuples $(\bm{e},q_{\tau})$, where $\bm{e}{=}\phi(x)$ is the task embedding and $q_{\tau}$ is the pheromone-deposition score associated with a verified tool-use trajectory that contains this edge.
Given a task $x$, we compute task-dependent pheromones by retrieval and similarity-weighted aggregation as in Eq.~\eqref{eq:dep}, estimate confidence by Eq.~\eqref{eq:confidence}, and then combine task-agnostic and task-dependent pheromones using Eq.~\eqref{eq:fuse-pher}.
We apply the same mechanism to tool-to-invocation edges for $\tau_{\text{Arg}}$.

\paragraph{Rationale and locality of the confidence score.}
The confidence score in Eq.~\eqref{eq:confidence} is computed \emph{locally} from the retrieved set $\mathcal{M}_{ij}(x)$ for the queried task $x$.
In particular, $n_{\text{act}}$ counts the number of retrieved memories that pass activation thresholds, $s_{\max}$ is the maximum similarity among activated memories, and $\bar{q}_{\tau}$ is the mean pheromone-deposition score over the same activated subset.
We incorporate $\bar{q}_{\tau}$ in confidence to gate task-dependent pheromones when retrieval returns semantically similar but consistently low-quality memories.
This design keeps the fused explicit transition prior conservative, assigning higher weight to context-aware pheromones only when retrieved evidence is both relevant (high similarity) and reliable (high average quality), and otherwise falling back to task-agnostic pheromones.

\subsection{Reward signals and pheromone-deposition score}
\label{app:reward-structure}

We distinguish two quantities computed per rollout: the episode return $R$ used for policy optimization, and the pheromone-deposition score $q_{\tau}(\xi)$ used for pheromone deposition and memory writing.
They are computed together during rollout generation but serve different roles in training.

\paragraph{Reward structure and episode return.}
For each rollout, we compute an episode return $R$ as the sum of step-level rewards and an outcome-level bonus.
At each step $t$, the step reward decomposes into an intent component and an execution component,
\begin{equation}
r_t = r_t^{\text{intent}} + r_t^{\text{exec}}.
\end{equation}
The intent reward $r_t^{\text{intent}}$ encourages selecting the correct next base tool.
Concretely, we assign $+0.5$ for an exact match to the reference next base tool, and $+0.2$ for a category-consistent selection (i.e., the predicted tool belongs to the same predefined tool category as the reference).
To mitigate error propagation, we additionally provide a recovery bonus of $+0.1$ when the previous step is incorrect but the current step returns to the reference tool.\footnote{This recovery term is only applied when $a_{t-1}\neq a^*_{t-1}$ and $a_t=a^*_t$.}
Otherwise, $r_t^{\text{intent}}=0$.
The execution reward $r_t^{\text{exec}}$ is assigned after simulating the selected tool invocation (Sec.~\ref{app:tool-sim}).
We assign $+0.5$ when the simulator returns a valid execution without triggering an error pattern, and $-0.5$ when an execution error or timeout is detected.
We treat invalid argument formatting as an execution failure and apply the same penalty.

After the rollout terminates, we add an outcome-level bonus that rewards completed episodes and, when completion is not achieved, provides graded bonuses based on the final Match Ratio.
The episode return used by GRPO is
\begin{equation}
R = \sum_{t=1}^{T} r_t \;+\; r_{\text{outcome}}.
\label{eq:return}
\end{equation}

\paragraph{Pheromone-deposition score.}
The pheromone deposition mechanism in Eq.~(1) relies on a pheromone-deposition score $q_{\tau}(\xi)\in[0,1]$.
In our benchmark setting, each instance provides a reference tool-use trajectory $\xi^\ast = (a_1^\ast,\dots,a_T^\ast)$.
Given a rollout trajectory $\xi=(a_1,\dots,a_T)$, we define
\begin{equation}
q_{\tau}(\xi) \;=\; \mathrm{Match}(\xi,\xi^\ast)
\;=\;
\frac{1}{T}\sum_{t=1}^{T} \mathbb{I}\!\left[\mathrm{tool}(a_t)=\mathrm{tool}(a_t^\ast)\right],
\label{eq:q-match}
\end{equation}
where $\mathrm{tool}(\cdot)$ extracts the base tool identity from the two-layer action, and the indicator compares step-aligned tool identities.
This definition uses the reference tool-use trajectory, and it is computed independently from the environmental return $R$ in Eq.~\eqref{eq:return}.
We maintain both quantities since $R$ provides the learning signal for policy optimization, while $q_{\tau}(\xi)$ only updates pheromone statistics.

\paragraph{Teacher forcing and model-driven quality.}
During mixed rollouts with oracle guidance, we record a per-step teacher-forcing mask indicating whether $a_t$ was oracle-selected.
We additionally report a model-driven Match Ratio computed on the subset of steps not teacher-forced:
\begin{equation}
q_{\text{model}}(\xi)
=
\frac{1}{|\mathcal{T}_{\text{model}}|}\sum_{t\in\mathcal{T}_{\text{model}}}
\mathbb{I}\!\left[\mathrm{tool}(a_t)=\mathrm{tool}(a_t^\ast)\right],
\qquad
\mathcal{T}_{\text{model}}=\{t:\; a_t \text{ is not oracle-selected}\}.
\label{eq:q-model}
\end{equation}
This quantity diagnoses how much alignment is attributable to autonomous decisions.
In addition, we only collect elite replay trajectories after $p_{\text{tf}}$ decays below a threshold, which limits the impact of heavily teacher-forced trajectories on subsequent pheromone reinforcement.

\subsection{Tool execution simulation}
\label{app:tool-sim}

Our training environment executes tool calls through a simulator, not by invoking external APIs.
This design keeps rollouts reproducible, avoids network variability, and prevents unintended side effects.
Given the selected tool and its argument invocation, the simulator produces a tool output and a completion signal that indicates whether the episode has reached a terminal success state.
The simulator can replay cached outputs from an offline database, and it also supports a pluggable LLM-based simulator backend that generates tool outputs via an OpenAI-compatible interface.
The execution reward in Sec.~\ref{app:reward-structure} is computed from this simulated output, assigning positive reward to valid executions and penalizing execution errors or malformed outputs.
This simulator-based reward is used for training and rollout feedback. For evaluation, Task Success Rate is measured by GPT-4o as an outcome judge under a fixed protocol, based on the task requirement, generated tool-use trajectory, and final tool outputs. The judge evaluates task completion rather than reference-order matching.

\paragraph{Fairness of comparisons.}
To ensure fair comparisons, all methods share the same backbones, tool set, decoding configuration, rollout horizon limits, and evaluation protocol, and they are trained under the same computational budget whenever applicable.
For baseline methods that require tool-format awareness (e.g., producing valid tool names or argument fields), we apply the same lightweight tool-format fine-tuning procedure used in our approach (e.g., the same LoRA setup on identical tool-call supervision) to calibrate tool syntax and reduce avoidable execution failures.
This control isolates the effect of the learning algorithm and sampling strategy, and it prevents differences in tool-call formatting from dominating the reported performance.

\section{Datasets}
\label{app:datasets}

We evaluate on three long-horizon tool-use benchmarks: Toolathlon, TRAJECT-Bench, and TOUCAN.
Each benchmark provides (i) executable tools or tool interfaces, and (ii) a reference tool-use sequence, which enables fine-grained comparison between predicted tool-use sequences and the reference.
Following our main experimental protocol, we use the official splits when available, and we only keep episodes whose reference tool-use sequences are verifiably successful.

To align with our Tool-Transition Graph action space (Sec.~\ref{sec:tool-graph}), we report dataset-scale statistics in terms of:
\textbf{L1 Tools}, the number of base tools in $\mathcal{V}_{\text{Tool}}$;
\textbf{L2 Tools}, the number of argument-invocation patterns aggregated across tools in $\mathcal{V}_{\text{Arg}}$;
and \textbf{Avg Steps}, the average number of tool calls per verifiably successful reference tool-use sequence in the retained episodes.

\paragraph{Toolathlon.}
Toolathlon is a long-horizon benchmark designed for realistic, execution-based evaluation across diverse software applications.
After mapping tool calls into our Tool-Transition Graph action representation, Toolathlon contains 618 L1 tools and 1{,}250 L2 argument-invocation patterns.
Among the retained verifiably successful reference sequences, the average tool-call length is 13.48 steps.

\paragraph{TRAJECT-Bench.}
TRAJECT-Bench is a trajectory-aware benchmark that emphasizes correctness of tool selection, tool arguments, and dependency or ordering constraints along the full tool-use trajectory.
In our Tool-Transition Graph representation, TRAJECT-Bench contains 381 L1 tools and 715 L2 argument-invocation patterns.
Among the retained verifiably successful reference sequences, the average tool-call length is 6.36 steps.

\paragraph{TOUCAN (Multi-step only).}
TOUCAN is a large-scale tool-agent dataset synthesized from real-world Model Context Protocol (MCP) environments.
We only consider the Multi-step subset, defined as instances with at least two tool calls in the reference sequence.
In our Tool-Transition Graph representation, the Multi-step subset contains 850 L1 tools and 15{,}294 L2 argument-invocation patterns.
Among the retained verifiably successful reference sequences, the average tool-call length is 3.16 steps.

\begin{table*}[ht]
\centering
\small
\caption{\textbf{Dataset Statistics.} We summarize the scale of the three benchmarks under our Tool-Transition Graph action representation.
L1 Tools denotes the number of base tools, and L2 Tools denotes the number of argument-invocation patterns.
Avg Steps reports the average number of tool calls per verifiably successful reference tool-use sequence in the retained episodes.
For TOUCAN, we only consider Multi-step instances.}
\label{tab:dataset_stats}
\setlength{\tabcolsep}{6pt}
\begin{tabular}{lccccc}
\toprule
Dataset & Tool source & L1 Tools & L2 Tools & \#Instances & Avg Steps \\
\midrule
Toolathlon & MCP apps & 618 & 1{,}250 & 108 tasks & 13.48 \\
TRAJECT-Bench & executable APIs & 381 & 715 & 5{,}870 queries & 6.36 \\
TOUCAN (Multi-step) & MCP servers & 850 & 15{,}294 & 1{,}646{,}546 traj. & 3.16 \\
\bottomrule
\end{tabular}
\end{table*}

\section{GPT-4o Judge Protocol for Task Success Rate}
\label{app:gpt4o_judge}

Task Success Rate (TSR) measures whether a generated tool-use trajectory successfully completes the target task.
We evaluate TSR using GPT-4o as an outcome judge under a fixed evaluation protocol \citep{mateval, kgerror, c3tg}.
This evaluation is used only for reporting test performance and is not used during training or pheromone updating.

For each test instance, we provide the judge with the task instruction, the generated tool-use trajectory, and the final tool outputs.
We do not provide the method name to the judge.
The judge is asked to evaluate whether the final outcome satisfies the task requirement, independent of whether the generated trajectory follows the reference ordering.
This design is important because a task can sometimes be completed through a valid alternative tool-use path even when its Match Ratio to the reference trajectory is not perfect.

The judge outputs a binary decision, where \texttt{success} indicates that the generated trajectory completes the task and \texttt{failure} indicates that it does not.
A trajectory is labeled as successful only when the final output satisfies the main task requirement and is supported by the executed tool outputs.
Partial completion, missing outputs, invalid tool results, hallucinated results, or outputs that do not address the requested task are labeled as failures.
Minor formatting differences are ignored when they do not affect task correctness.
We use the same judge prompt and deterministic decoding setting for all methods.

Formally, for a test set with $N$ instances, let $y_i \in \{0,1\}$ denote the GPT-4o judge label for the $i$-th generated trajectory, where $1$ indicates success.
We compute TSR as
\begin{equation}
\mathrm{TSR} = \frac{1}{N}\sum_{i=1}^{N} y_i \times 100\%.
\end{equation}

The judge prompt template is shown below.

\begin{quote}
\small
You are an impartial evaluator for long-horizon tool-use agents.  
Given a task instruction, a generated tool-use trajectory, and the final tool outputs, determine whether the agent successfully completed the task.

Evaluate only the final task outcome.  
Do not require the generated trajectory to follow the reference tool ordering.  
A different tool-use path can be marked as successful if it produces a correct final outcome.  
Mark the trajectory as failure if the final output is missing, incomplete, unsupported by tool outputs, based on invalid tool execution, or does not satisfy the task requirement.

Return your judgment in the following format:

\texttt{\{}  
\texttt{"success": 0 or 1,}  
\texttt{"reason": "brief explanation"}  
\texttt{\}}
\end{quote}

\section{Ablation Study on Pheromone Influence Parameter $\beta$}
\label{appendix:beta_ablation}

This appendix studies the pheromone influence parameter $\beta$ in Eq.~\eqref{eq:sampling}, which scales the explicit transition prior during pheromone-guided sampling. $\beta$ controls how strongly historically verified tool-transition patterns bias the next tool choice, shaping the exploration exploitation trade-off in long-horizon planning.

\begin{figure*}[ht]
    \centering
    \includegraphics[width=\textwidth]{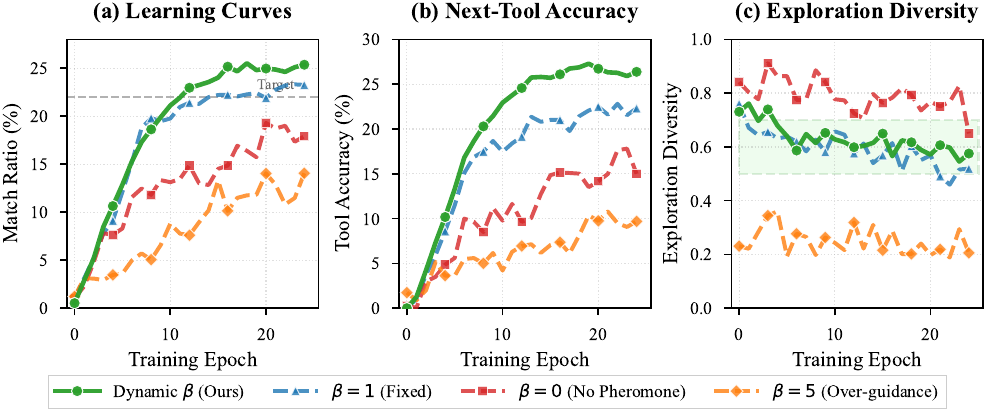}
    \caption{\textbf{Ablation study of pheromone influence parameter $\beta$ on Toolathlon benchmark.} 
    (a)~\textbf{Learning curves} show that dynamic $\beta$ annealing achieves the highest Match Ratio (25.25\%), significantly outperforming fixed strategies. $\beta=0$ (no pheromone) exhibits slow, noisy convergence, while $\beta=5$ (over-guidance) becomes trapped in local optima. 
    (b)~\textbf{Next-tool accuracy} mirrors learning trends, with dynamic $\beta$ reaching 27.16\% through effective balance of exploration and exploitation. 
    (c)~\textbf{Exploration diversity} exhibits natural fluctuations due to finite-sample estimation from rollouts, but reveals clear trends: $\beta=0$ maintains excessively high diversity (unfocused exploration), $\beta=5$ shows extremely low diversity (trapped exploitation), while dynamic $\beta$ demonstrates an optimal trajectory—starting with high exploration (0.75) and smoothly transitioning into the optimal range (0.55-0.65) that balances verified pattern reuse with adaptive exploration.}
    \label{fig:beta_ablation}
\end{figure*}

Specifically, we sample tools from
\begin{equation}
p(a_t \mid s_t, a_{t-1}) \propto \exp\!\Big(\log \pi_\theta(a_t \mid s_t) + \beta \log \tau_{\text{Tool}}(a_{t-1}, a_t \mid x)\Big),
\end{equation}
where $\beta \ge 0$ determines the strength of pheromone guidance. $\beta=0$ removes pheromone, while larger $\beta$ increasingly favors transitions with higher pheromone, which can reduce exploration if set too aggressively.

\subsection{Experimental Setup}
We ablate $\beta$ on Toolathlon with Qwen2.5-7B. We compare: (i) $\beta=0$ (no pheromone), (ii) $\beta=1$ (fixed moderate), (iii) $\beta=5$ (strong over-guidance), and (iv) our dynamic schedule, where $\beta$ is linearly increased from $0$ to $0.8$ over the first 30\% of epochs and then kept at $0.8$. All variants share the same backbone, optimizer, learning rate schedule, rollout group size ($M=5$), and evaluation protocol. We train for 25 epochs and report Match Ratio (Match R.) and Next-tool Accuracy (TAcc). We also measure \emph{Exploration Diversity} as the normalized entropy of the action distribution over tools:
\begin{equation}
\text{Diversity} = \frac{H(\pi)}{\log |A|} \in [0,1].
\end{equation}

\subsection{Results}
Table~\ref{tab:beta_ablation} shows that pheromone guidance improves performance over $\beta=0$, while excessive guidance ($\beta=5$) harms both Match R.\ and TAcc. Our dynamic schedule achieves the best results, indicating that gradually increasing pheromone influence matches the increasing reliability of accumulated pheromone statistics.

\begin{table}[ht]
\centering
\small
\caption{Ablation on $\beta$ for Toolathlon (Qwen2.5-7B). We report Match Ratio (Match R., \%), Next-tool Accuracy (TAcc, \%), and final Exploration Diversity.}
\label{tab:beta_ablation}
\setlength{\tabcolsep}{6pt}
\begin{tabular}{lccc}
\toprule
\textbf{Setting} & \textbf{Match R.} & \textbf{TAcc} & \textbf{Final Diversity} \\
\midrule
$\beta = 0$ (No pheromone) & 21.23 & 19.86 & 0.76 \\
$\beta = 1$ (Fixed) & 23.93 & 22.05 & 0.53 \\
$\beta = 5$ (Over-guidance) & 17.80 & 16.50 & 0.25 \\
\textbf{Dynamic $\beta$ (Ours)} & \textbf{25.25} & \textbf{27.16} & 0.58 \\
\bottomrule
\end{tabular}
\end{table}

\paragraph{Interpretation.}
With $\beta=0$, the policy remains highly diverse but learns slowly, suggesting unfocused exploration. With $\beta=5$, the policy becomes nearly deterministic and underperforms, consistent with premature over-exploitation of early transition patterns. Fixed $\beta=1$ provides a stable improvement over $\beta=0$, but it cannot adapt to the evolving quality of pheromone statistics. Dynamic $\beta$ preserves early diversity for discovery and then reduces diversity to a stable mid-range as training progresses, aligning improved Match R.\ with higher TAcc. Variance in the diversity curve is expected because it is estimated from a small number of rollouts per epoch ($M=5$), while Match R.\ and TAcc average across trajectories and evaluation instances.

Pheromone guidance benefits long-horizon tool planning, but its strength must be calibrated. Extreme settings either remove reusable transition information ($\beta=0$) or suppress exploration ($\beta=5$). A progressive schedule for $\beta$ yields the best balance, improving both sequence-level Match R.\ and step-level TAcc.

\section{Training Dynamics and Pheromone Evolution}
\label{app:dynamics}

\begin{figure}[ht]
\centering
\includegraphics[width=0.75\linewidth]{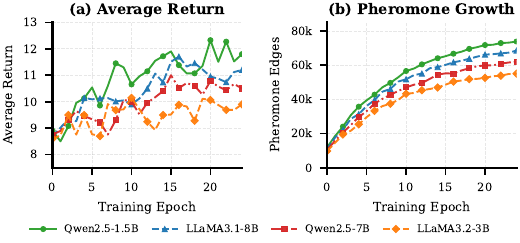}
\caption{\textbf{Training Dynamics of PhGPO.} (a) \textbf{Average Return}: All backbone models show a consistent upward trend in average return, indicating stable policy improvement during training. (b) \textbf{Pheromone Graph Growth}: The number of discovered edges grows rapidly in early epochs and stabilizes later, reflecting fast discovery of feasible tool transitions followed by refinement on a stable set of edges.}
\label{fig:dynamics}
\end{figure}

\noindent\textbf{Training dynamics.}
To assess optimization stability, we visualize the evolution of policy performance together with the evolution of the pheromone structure in Figure~\ref{fig:dynamics}. We report the \emph{average return} measured over training rollouts for each backbone. Figure~\ref{fig:dynamics}(a) shows a consistent upward trend across all four backbone models, indicating that PhGPO steadily improves the rollout-level objective. The smallest backbone, Qwen2.5-1.5B, reaches competitive returns compared with larger models, which suggests that the reusable
tool-transition patterns reduces the difficulty of long-horizon exploration for lower-capacity policies. The curves exhibit sustained improvement without abrupt collapses, which is consistent with stable optimization under our training configuration.

\noindent\textbf{Pheromone graph growth.}
Figure~\ref{fig:dynamics}(b) tracks the growth of the pheromone graph, measured by the number of discovered edges that receive pheromone statistics. These edges correspond to tool-transition edges, and they can also include argument-invocation edges when tool invocations are represented at the argument level. We observe rapid edge discovery during early epochs, followed by a clear saturation. This trend indicates that the agent quickly identifies a broad set of feasible tool transitions and then concentrates updates on a more stable subset. Under pheromone evaporation, older statistics gradually lose influence, which keeps the explicit transition prior responsive to recent verified tool-use trajectories while preventing unbounded accumulation.

\noindent\textbf{Relationship between returns and pheromone.}
The increase in average return coincides with the period where the pheromone graph expands and then stabilizes. This alignment suggests that improving performance is accompanied by accumulating transition evidence on the pheromone graph. Once most frequently used transitions have been identified, performance improvements continue primarily through adjusting the relative strengths of already discovered edges and improving the policy probabilities on those transitions.

\section{Hyperparameter Sensitivity Analysis}
\label{appendix:hyperparameter}

\begin{figure*}[t]
    \centering
    \includegraphics[width=\textwidth]{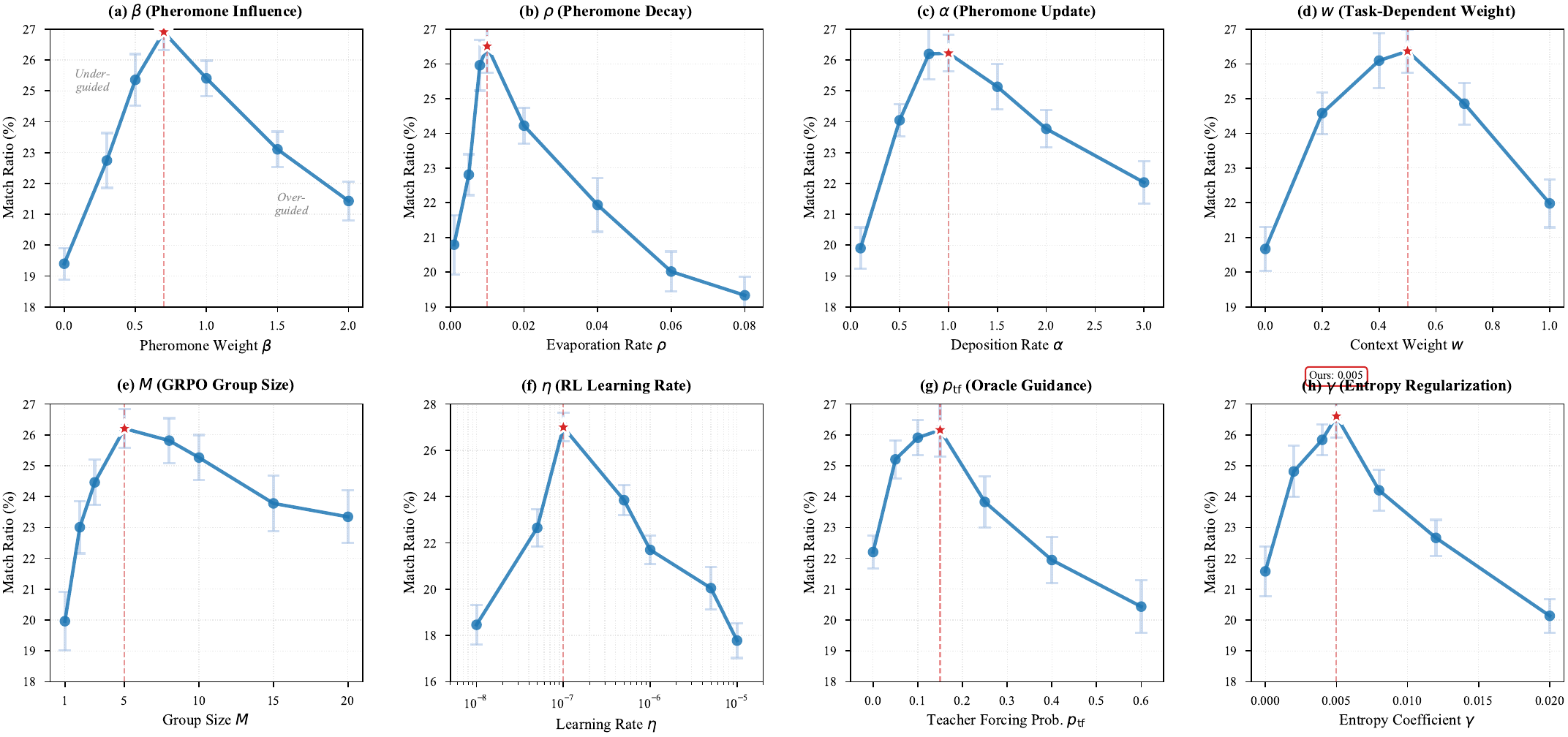}
    \caption{\textbf{Hyperparameter sensitivity analysis on Toolathlon.} We evaluate 8 hyperparameters for pheromone guidance and policy optimization. Red stars indicate our selected values; shaded regions show $\pm 1$ standard deviation across 3 seeds. \textbf{Row 1:} (a)~$\beta_{\max}$; (b)~$\rho$; (c)~$\alpha$; (d)~$w$. \textbf{Row 2:} (e)~$M$; (f)~$\eta$; (g)~$p_{\mathrm{tf}}$; (h)~$\gamma$.}
    \label{fig:hyperparameter_8}
\end{figure*}

This appendix reports a sensitivity analysis of eight key hyperparameters in PhGPO on Toolathlon.
We vary one hyperparameter at a time while keeping all others fixed to the default values in Table~\ref{tab:hyperparameters_8}.
We train with Qwen2.5-7B and report Match Ratio on the test set as a fine-grained sequence-level diagnostic.
The main experiments report both Match Ratio and Task Success Rate, while this sensitivity analysis uses Match Ratio because it provides a continuous signal for comparing hyperparameter trends.
Each configuration is evaluated with 3 random seeds, and Figure~\ref{fig:hyperparameter_8} shows the mean with $\pm 1$ standard deviation.
Red stars and dashed lines indicate our chosen values.

\noindent\textbf{Hyperparameters.}
We group the parameters into pheromone guidance and policy optimization.
Pheromone guidance includes the maximum pheromone influence $\beta_{\max}$ in Eq.~\eqref{eq:sampling}, evaporation rate $\rho$ and deposition rate $\alpha$ in Eq.~\eqref{eq:agn}, and the task-dependent pheromone weight $w$ in Eq.~\eqref{eq:fuse-pher}.
Policy optimization includes GRPO group size $M$, RL fine-tuning learning rate $\eta$, final teacher forcing probability $p_{\mathrm{tf}}$, and entropy regularization coefficient $\gamma$.

\begin{table}[ht]
\centering
\caption{Key hyperparameters in PhGPO and selected values.}
\label{tab:hyperparameters_8}
\scriptsize
\setlength{\tabcolsep}{6pt}
\begin{tabular}{llcp{5.5cm}}
\toprule
\textbf{Category} & \textbf{Parameter} & \textbf{Value} & \textbf{Description} \\
\midrule
\multirow{4}{*}{\textbf{Pheromone}} 
& $\beta_{\max}$ & 0.7 & Maximum pheromone influence weight in Eq.~\eqref{eq:sampling} \\
& $\rho$ & 0.01 & Evaporation rate in Eq.~\eqref{eq:agn} \\
& $\alpha$ & 1.0 & Deposition rate in Eq.~\eqref{eq:agn} \\
& $w$ & 0.5 & Task-dependent pheromone weight in Eq.~\eqref{eq:fuse-pher} \\
\midrule
\multirow{4}{*}{\textbf{Optimization}}
& $M$ & 5 & GRPO group size, measured as rollouts per instance \\
& $\eta$ & $10^{-7}$ & Learning rate for RL fine-tuning \\
& $p_{\mathrm{tf}}$ & 0.15 & Final teacher forcing probability \\
& $\gamma$ & 0.005 & Entropy regularization coefficient \\
\bottomrule
\end{tabular}
\end{table}

\noindent\textbf{Overall observations.}
Figure~\ref{fig:hyperparameter_8} shows that our defaults fall within, or very close to, the best-performing regions across all eight parameters.
Among pheromone parameters, $\beta_{\max}$ and $\rho$ are the most sensitive.
A small $\beta_{\max}$ weakens the influence of the explicit transition prior, while an overly large $\beta_{\max}$ can suppress exploration.
The evaporation rate $\rho$ performs best around $0.01$, where older statistics decay slowly enough to remain useful but not so slowly that early transitions dominate.
In contrast, $\alpha$ is comparatively robust, with a broad near-optimal region around $1.0$.
The task-dependent pheromone weight $w$ peaks around $0.5$, indicating that combining task-dependent and task-agnostic pheromones is more effective than relying on either source alone.

For optimization parameters, $M$ shows diminishing returns.
Performance improves rapidly from $M=1$ and saturates around $M=5$.
The RL learning rate $\eta$ has a narrow optimum on a log scale, consistent with the need for conservative fine-tuning in the final stage.
The final teacher forcing probability $p_{\mathrm{tf}}$ follows a U-shaped pattern, where too little oracle guidance destabilizes early pheromone accumulation and too much oracle guidance limits autonomous improvement.
The entropy coefficient $\gamma$ has a broad plateau around $0.005$, maintaining exploration without preventing convergence.

\noindent\textbf{Practical guidance.}
When adapting PhGPO, tuning $\rho$ and $\beta_{\max}$ is most important because they directly control the strength and persistence of pheromone guidance.
The remaining parameters have wider stable regions, and our defaults provide a strong starting point.
Progressive schedules for $\beta$, $w$, and $p_{\mathrm{tf}}$ are consistent with these sensitivity trends because they avoid applying strong pheromone guidance before the corresponding statistics become reliable.

\section{Generality Across Policy Optimization Backbones}
\label{app:optimizers}

This appendix shows that PhGPO is not tied to a specific policy optimization backbone.
PhGPO defines (i) pheromone as an explicit transition prior on tool-transition edges and tool-to-invocation edges (Sec.~\ref{sec:pheromone}), (ii) pheromone-guided sampling during rollout generation (Eq.~\eqref{eq:sampling}), and (iii) pheromone updates via deposition and evaporation (Eq.~\eqref{eq:agn} and Eq.~\eqref{eq:fuse-pher}).
These components are independent of the choice of the policy-gradient optimizer used to update $\pi_\theta$.
In practice, we can replace GRPO with other on-policy optimizers, while keeping the same pheromone-guided rollout distribution and pheromone bookkeeping.

\subsection{GRPO, PPO, and RLOO backbones}
We describe three interchangeable optimization backbones under the same rollout protocol.
For each task instance, we sample $M$ tool-use trajectories from the pheromone-guided distribution in Eq.~\eqref{eq:sampling}, obtain sequence-level returns $\{R_m\}_{m=1}^{M}$, and update the policy using a clipped policy-gradient objective.
Across all three backbones, pheromone-guided sampling and pheromone updates remain unchanged.

\noindent\textbf{GRPO (Group Relative Policy Optimization).}
GRPO uses group-relative normalization to construct a sequence-level advantage for each rollout:
\begin{equation}
A_m = \frac{R_m - \bar{R}}{\sigma_R + \epsilon},
\qquad
\bar{R}=\frac{1}{M}\sum_{m=1}^{M}R_m,
\label{eq:grpo_adv_app}
\end{equation}
and applies a PPO-style clipped ratio objective with $A_m$ (Sec.~\ref{sec:training}).
This backbone does not require an explicit value function and leverages within-group contrasts.

\noindent\textbf{PPO (Proximal Policy Optimization).}
PPO constructs an advantage using a baseline $b(\cdot)$, which can be implemented as a learned value head or a running baseline.
In our setting, we use a sequence-level advantage
\begin{equation}
A_m = R_m - b(x),
\label{eq:ppo_adv_app}
\end{equation}
and optimize the standard clipped surrogate objective:
\begin{equation}
\mathcal{L}^{\text{PPO}}
=
-\mathbb{E}\Big[\min\big(r_t(\theta)A_m,\ \mathrm{clip}(r_t(\theta),1-\varepsilon,1+\varepsilon)A_m\big)\Big],
\label{eq:ppo_obj_app}
\end{equation}
where $r_t(\theta)=\pi_\theta(a_t\mid s_t)/\pi_{\theta_{\text{old}}}(a_t\mid s_t)$ is the probability ratio at tool-decision steps.

\noindent\textbf{RLOO (Leave-one-out baseline).}
RLOO avoids learning a value function and uses a leave-one-out baseline computed from the other rollouts in the same group:
\begin{equation}
b_{-m}=\frac{1}{M-1}\sum_{j\ne m} R_j,
\qquad
A_m = R_m - b_{-m}.
\label{eq:rloo_adv_app}
\end{equation}
The policy is then updated using the same clipped objective as Eq.~\eqref{eq:ppo_obj_app}, with $A_m$ replaced by the leave-one-out advantage.

\subsection{Experimental protocol on Toolathlon}
We evaluate optimizer-agnosticity on Toolathlon using Qwen2.5-7B and Llama3.1-8B.
All runs use the same action space, pheromone-guided sampling rule (Eq.~\eqref{eq:sampling}), pheromone update rule (Eq.~\eqref{eq:agn}), and the same progressive pipeline in Sec.~\ref{sec:training}.
The only change is the policy optimization backbone used to update $\pi_\theta$.
We report sequence-level Match Ratio (Match R., \%) and Next-tool Accuracy (TAcc, \%).

\begin{table}[ht]
\centering
\small
\caption{\textbf{PhGPO with different policy optimization backbones on Toolathlon.}
For each backbone optimizer, we report results with pheromone-guided sampling (Eq.~\eqref{eq:sampling}) and without pheromone guidance by setting $\beta{=}0$ in sampling.
Higher is better ($\uparrow$).}
\label{tab:optimizer_agnostic}
\setlength{\tabcolsep}{6pt}
\begin{tabular}{lcccc}
\toprule
\multirow{2}{*}{\textbf{Optimizer variant}}
& \multicolumn{2}{c}{\textbf{Qwen2.5-7B}}
& \multicolumn{2}{c}{\textbf{Llama3.1-8B}} \\
\cmidrule(lr){2-3}\cmidrule(lr){4-5}
& Match R. $\uparrow$ & TAcc $\uparrow$ & Match R. $\uparrow$ & TAcc $\uparrow$ \\
\midrule

\textbf{PhGPO + GRPO} & 25.25 & 27.16 & 22.83 & 23.09 \\
\quad w/o pheromone guidance ($\beta{=}0$) & 21.23 & 19.86 & 18.75 & 16.45 \\
\midrule

\textbf{PhGPO + PPO} & 23.27 & 25.03 & 18.21 & 24.05 \\
\quad w/o pheromone guidance ($\beta{=}0$) & 20.78 & 23.31 & 15.72 & 21.22 \\
\midrule

\textbf{PhGPO + RLOO} & 20.21 & 24.47 & 18.51 & 22.69 \\
\quad w/o pheromone guidance ($\beta{=}0$) & 18.61 & 23.26 & 17.39 & 21.51 \\

\bottomrule
\end{tabular}
\end{table}

\begin{figure*}[ht]
\centering
\includegraphics[width=\textwidth]{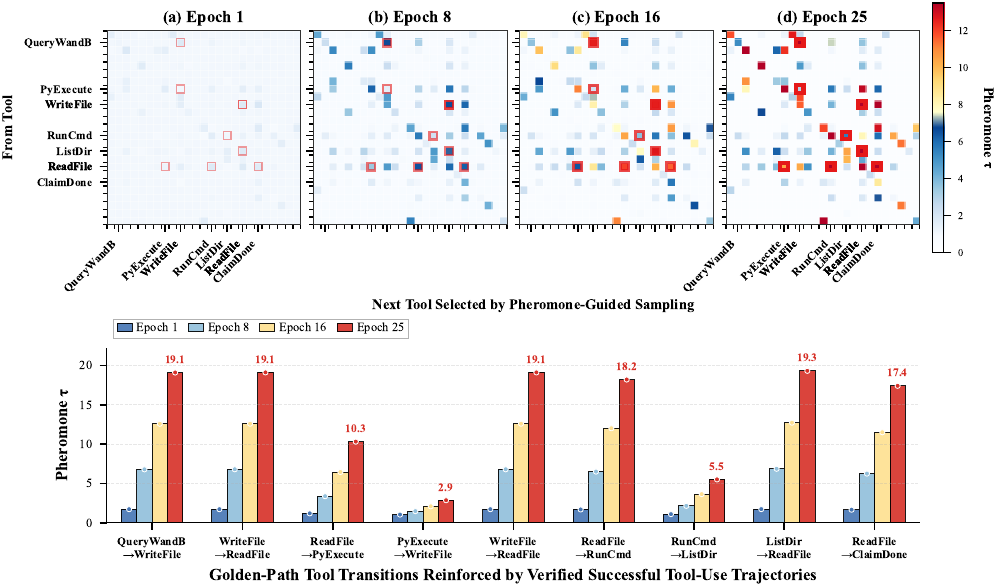}
\caption{\textbf{Pheromone evolution on a longer tool-use trajectory, visualized on a 10-tool subset.}
Top: fused tool-transition pheromone $\tau_{\text{Tool}}(\cdot\mid x)$ restricted to ten tools at checkpoint, with reference-chain edges highlighted in red.
Bottom: pheromone values on the highlighted reference-chain transitions at the same checkpoint.
The visualization shows that pheromone concentrates on reference-chain transitions while off-chain transitions fade under evaporation, indicating that the explicit transition prior remains selective on longer tool-use trajectories.}
\label{fig:pheromone_10tools}
\end{figure*}

\section{Extended Visualization on Longer Tool-use Trajectories}
\label{app:long_horizon}

This appendix provides an extended visualization showing that the pheromone-based explicit transition prior remains informative on longer tool-use trajectories.
For readability, we visualize a \emph{10-tool subset} of the tool-transition space, while the underlying episode involves a longer reference tool-use trajectory in the same MCP execution environment.
All visualizations report the \emph{fused} tool-transition pheromone $\tau_{\text{Tool}}(\cdot\mid x)$ defined in Eq.~\eqref{eq:fuse-pher}, where task-dependent evidence is retrieved by Eq.~\eqref{eq:dep} and confidence is computed by Eq.~\eqref{eq:confidence}.
Pheromone updates follow deposition and evaporation in Eq.~\eqref{eq:agn}.

\paragraph{10-tool subset visualization}
\label{app:long_horizon_10tools}

Figure~\ref{fig:pheromone_10tools} visualizes pheromone evolution on a representative long-horizon instance using the following 10 tools:
\begin{center}
\small
\texttt{QueryWandB $\rightarrow$ WriteFile $\rightarrow$ ReadFile $\rightarrow$ PyExecute $\rightarrow$ WriteFile $\rightarrow$ ReadFile $\rightarrow$ RunCmd $\rightarrow$ ListDir $\rightarrow$ ReadFile $\rightarrow$ ClaimDone}
\end{center}

The heatmaps (top row) show the fused pheromone matrix restricted to these tools at several training checkpoints.
Reference-chain transitions that appear in this instance are highlighted in red.
As training proceeds, pheromone mass becomes increasingly concentrated on the reference-chain transitions within this 10-tool subset, while many off-chain transitions fade as evaporation reduces the influence of older or weakly supported statistics.
This pattern indicates that the explicit transition prior becomes more selective over time, amplifying transitions that recur in verified successful tool-use trajectories.

The bar plot (bottom row) reports pheromone values along the reference-chain transitions within the same 10-tool subset at the corresponding checkpoints.
Earlier transitions often exhibit stronger reinforcement because they are visited more frequently across verified successful trajectories and because early tool choices constrain later reachable states in long tool-use trajectories.
Later transitions remain distinguishable from the background, which indicates that the pheromone signal is not limited to prefix segments and can provide guidance throughout the chain.

\section{Time Efficiency: Path Discovery Speed}
\label{app:time_efficiency}

This appendix provides a qualitative analysis of training-time efficiency.
We compare PhGPO against an ablation that removes the pheromone prior, and examine how quickly each method discovers the first high-quality tool-use trajectory on Toolathlon episodes.
A trajectory is considered high-quality when its pheromone-deposition score satisfies $q_{\tau}(\xi)\ge 0.6$.

\noindent\textbf{Experimental protocol}
We evaluate two configurations on Toolathlon with Qwen2.5-7B:

\begin{itemize}
  \item \textbf{PhGPO (Full)}: rollouts are sampled from the fused distribution that incorporates pheromone as an explicit transition prior (Eq.~\eqref{eq:sampling}), with the progressive schedule described in Sec.~\ref{sec:training}.
  \item \textbf{GRPO w/o pheromone}: we set $\beta=0$ throughout training, removing pheromone guidance during rollout generation while keeping the remaining pipeline identical (same curriculum, the same oracle-mixing rule via $p_{\text{tf}}$, and the same mixed objective).
\end{itemize}

For each Toolathlon episode, we record the earliest training step at which at least one generated trajectory reaches $q_{\tau}(\xi)\ge 0.6$, and summarize the trend across episodes and trajectory-length bins.

\begin{figure}[t]
\centering
\includegraphics[width=0.8\textwidth]{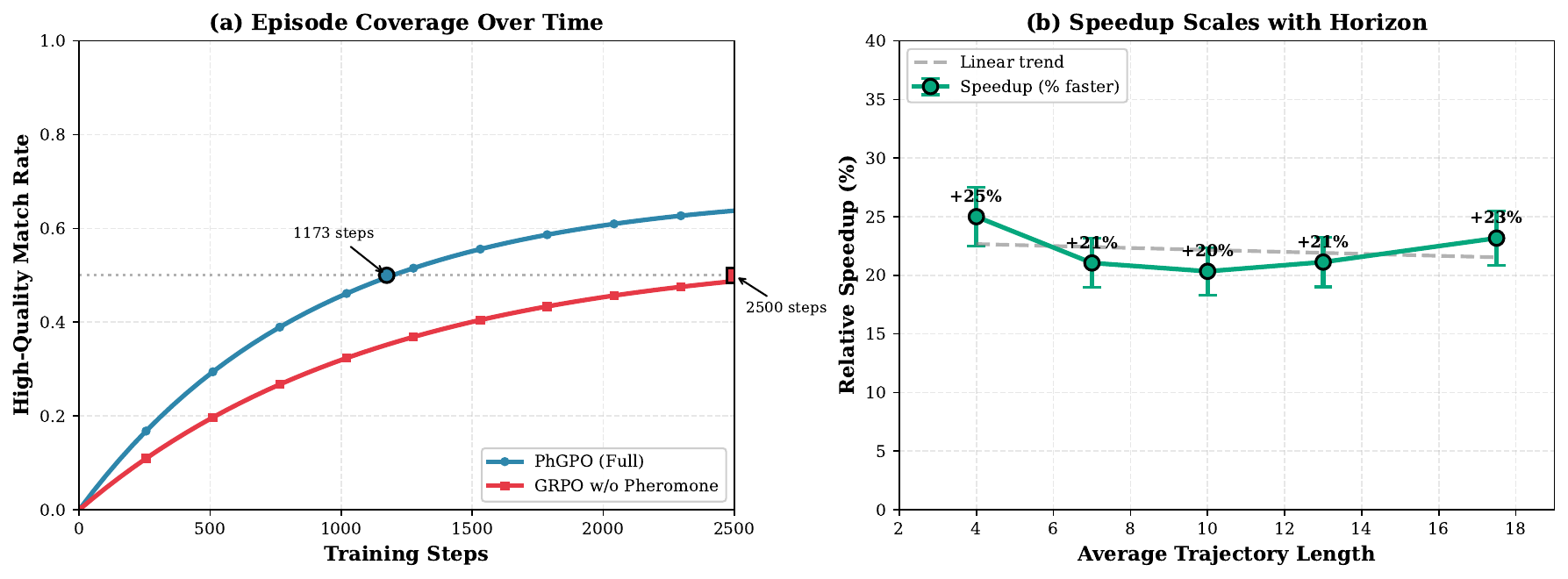}
\caption{\textbf{Time efficiency analysis on Toolathlon.}
We track the earliest training step at which a high-quality trajectory ($q_{\tau}(\xi)\ge 0.6$) is first generated for each episode.
PhGPO reaches broad coverage earlier than the ablation without pheromone, and the relative improvement remains broadly consistent across trajectory-length bins.}
\label{fig:time_efficiency}
\end{figure}

\noindent\textbf{Results.}
Figure~\ref{fig:time_efficiency} summarizes the path discovery speed measured by the first occurrence of a high-quality trajectory ($q_{\tau}(\xi)\ge 0.6$).
In panel (a), PhGPO maintains a higher high-quality match rate than the ablation without pheromone at matched training steps, and it reaches the same coverage level with fewer training steps.
The gap remains visible across the training range shown in the figure, indicating more efficient discovery of high-quality trajectories under the same budget.
Panel (b) reports the relative speedup across trajectory-length bins.
The speedup stays positive across bins and varies within a narrow range, and the fitted linear trend is close to flat.
This pattern suggests that the efficiency gain is not restricted to a specific horizon, and it remains broadly stable as the average trajectory length increases.

\noindent\textbf{Interpretation}
The trend is consistent with the role of pheromone as an explicit transition prior during trajectory generation.
Without pheromone ($\beta=0$), rollout groups sampled only from the policy can be dominated by uniformly low-quality trajectories in a large tool space, which weakens group-relative learning signals.
With pheromone guidance, historically verified tool-transition edges receive higher sampling probability, which reduces wasted exploration on low-potential transitions and increases the likelihood that each rollout group contains informative contrasts for group-based policy optimization.
This mechanism can benefit from partial matches: trajectories that reach $q_{\tau}(\xi)\ge 0.6$ indicate reusable transition evidence, and verified successful ones are used for pheromone updates, even when they are not exact matches to the reference.

\noindent\textbf{Implications}
These results indicate that PhGPO not only improves final performance, but also improves training-time efficiency by accelerating the discovery of high-quality tool-use trajectories.
This property is particularly relevant in execution-based environments where tool calls are costly.

\section{Simulated Tool Execution}
\label{app:simulated_tools}

Our training and evaluation use a \emph{simulated} tool environment.
Tool calls are not executed against real external APIs or services.
Instead, each tool invocation is routed to a tool-output simulator that returns a plausible response conditioned on the current context.
This design provides a controlled and reproducible feedback channel for learning tool-use trajectories, while avoiding external side effects and service availability issues.

\subsection{LLM-based tool-output simulator}
We implement an LLM-based simulator that generates tool outputs from structured context.
Given a tool variant name, its arguments, the task name, the user request, and a truncated history of recent tool interactions, the simulator constructs a prompt that includes:
(i) the task and user request,
(ii) the recent tool-call history (with any completion markers removed),
(iii) the current tool name, its original tool name, and its natural-language description,
(iv) the provided arguments (or expected keys when arguments are absent).
The simulator then queries an instruction-following model to produce a short response that resembles a real tool return value (for example, a JSON snippet, a status string, or a concise result summary).

In our experiments, we use \texttt{qwen3-235b-a22b-thinking-2507} as the simulator model through an OpenAI-compatible chat-completions endpoint.
We fix the simulator system prompt to enforce output-only behavior, and we strip common explanatory prefixes and markdown formatting during post-processing to keep outputs consistent with a tool API style.

\subsection{Completion detection via an explicit marker}
The simulator also provides a lightweight termination signal.
After generating a tool output, the simulator decides whether the current interaction history already completes the user request.
If it judges the task complete, it appends a dedicated marker \texttt{<<END>>} on a separate final line.
We remove this marker from the returned tool output and store the completion flag separately.
This enables consistent stopping behavior during rollout generation without requiring an additional model call.

\subsection{Caching and fallback behavior}
To reduce simulator calls and stabilize training, we cache simulator outputs using a content hash over the tool variant, task name, user request, truncated history, and arguments.
When the LLM simulator is unavailable or fails, we optionally fall back to a pool-based simulator that samples from pre-defined outputs, and we use a simple default template as a final fallback.
These mechanisms preserve the tool-use interface while keeping the environment feedback stable under large-scale training.

\section{Adaptation to Dynamic Tool Environments}
\label{app:dynamic_tools}

A practical advantage of PhGPO is that the transition prior is maintained explicitly on the MCP-grounded graph and its edge memories, instead of being absorbed only into policy parameters.
This design allows tool-environment changes to be handled through local graph and memory updates without global re-initialization.

When a new tool is introduced, we add the corresponding tool node, argument-invocation patterns, tool-transition edges, and tool-to-invocation edges.
The task-agnostic pheromone values on newly added edges are initialized to $\tau_0$, and their memory banks are initialized as empty.
If the new tool is not yet covered by updated reference trajectories or sampled rollouts, its associated edges remain at $\tau_0$ with empty memory banks.
Statistics start to accumulate only after subsequent trajectories traverse these edges.
In many practical settings, newly introduced tools are accompanied by argument schemas and a small number of updated demonstrations.
When such data are available, the same supervised warm-up can warm-start the corresponding tool and invocation edges without re-initializing the whole policy.

When a tool is deprecated, the corresponding node and edges can be immediately masked from the action space.
Existing pheromone values and memory entries for the deprecated tool are no longer queried during rollout generation.
Evaporation further reduces the influence of stale statistics over time.
If the tool identity remains but its API schema changes, the update mainly affects the argument-invocation patterns and the corresponding tool-to-invocation edges, while unaffected tool-transition edges can be preserved.

We validate this behavior with a tool-evolution test on Toolathlon using Qwen2.5-7B.
We merge several frequent atomic tool chains into macro tools that replace the original chains, simulating a realistic API upgrade.
Under this change, local graph and memory update recovers to about $87\%$ of the pre-change performance using about $16\%$ of the original RL training steps.
By comparison, global re-initialization requires about $38\%$ of the original RL training steps to reach the same recovery level.
Performance on unaffected tasks remains stable under local update.
These results indicate that the explicit graph-indexed transition prior helps PhGPO adapt to evolving tool environments with limited retraining.

\section{Scalability and Computational Overhead}
\label{app:overhead}

PhGPO adds three main components on top of GRPO: graph maintenance, pheromone-guided sampling, and task-dependent edge-memory retrieval.
These components are lightweight because pheromone values are indexed by graph edges and queried only for the current candidate set during rollout generation.
PhGPO does not require inference-time tree search, additional full-trajectory rollouts, or LLM-based judging during training.

We benchmark the end-to-end overhead of these components on Toolathlon with Qwen2.5-7B.
Table~\ref{tab:overhead} reports training-step time, peak GPU memory, and inference-step latency.
The full PhGPO pipeline uses $11.89$s per training step compared with $11.86$s for GRPO, and $59.3$ms per inference step compared with $58.5$ms for GRPO.
The peak GPU memory increases from $17.3$GB to $17.4$GB.
The incremental rows show that graph maintenance, pheromone-guided sampling, and edge-indexed retrieval each add only marginal overhead.
In practice, runtime remains dominated by rollout generation and the forward/backward computation of the base model.

\begin{table}[ht]
\centering
\scriptsize
\caption{\textbf{Runtime and memory overhead on Toolathlon.}
We report end-to-end overhead using Qwen2.5-7B.}
\label{tab:overhead}
\setlength{\tabcolsep}{5pt}
\begin{tabular}{lccccc}
\toprule
\textbf{Method} & \textbf{Train step (s)} & \textbf{Train overhead} & \textbf{Peak GPU (GB)} & \textbf{Infer/step (ms)} & \textbf{Infer overhead} \\
\midrule
GRPO & 11.86 & baseline & 17.3 & 58.5 & baseline \\
+ graph maintenance & 11.87 & +0.01s & 17.3 & 58.8 & +0.3ms \\
+ graph + pheromone sampling & 11.88 & +0.02s & 17.4 & 59.1 & +0.6ms \\
Full PhGPO & 11.89 & +0.03s & 17.4 & 59.3 & +0.8ms \\
\bottomrule
\end{tabular}
\end{table}

The small additional cost comes from local operations on a sparse tool-transition graph.
Graph maintenance updates only traversed edges, pheromone-guided sampling reweights the Top-$K$ candidate set, and task-dependent retrieval is performed over edge-indexed memory banks.
These results suggest that PhGPO scales favorably in large tool spaces because it avoids global search over all possible tool sequences.

\section{Random Exploration and Pheromone Guidance}
\label{app:random_vs_pheromone}

Both random exploration and pheromone guidance affect the exploration-exploitation trade-off, but they play different roles.
The random exploration probability $p_{\mathrm{rand}}$ introduces a small memoryless perturbation at the current decision step.
In contrast, pheromone is a history-informed explicit transition prior accumulated across episodes.
It biases rollout generation toward tool-transition and argument-invocation patterns that repeatedly appear in verified successful trajectories.
Therefore, $p_{\mathrm{rand}}$ provides undirected local exploration, while pheromone provides directed cross-episode exploration.

To quantify their relative effects, we compare four settings on Toolathlon with Qwen2.5-7B.
The results are shown in Table~\ref{tab:random_vs_pheromone}.
Removing pheromone causes the largest drop in Match Ratio and Next-tool Accuracy while keeping diversity high, which is consistent with unfocused exploration.
Removing random exploration causes only a small decrease, indicating that pheromone is the primary source of effective exploration guidance.
Increasing $p_{\mathrm{rand}}$ from $0.05$ to $0.10$ increases diversity but slightly hurts performance, suggesting that stronger random perturbation can dilute the explicit transition prior.

\begin{table}[t]
\centering
\small
\caption{\textbf{Random exploration and pheromone guidance on Toolathlon.}
We compare the effects of removing pheromone guidance and changing the random exploration probability $p_{\mathrm{rand}}$ using Qwen2.5-7B.}
\label{tab:random_vs_pheromone}
\setlength{\tabcolsep}{6pt}
\begin{tabular}{lccccc}
\toprule
\textbf{Setting} & \textbf{$\beta$ schedule} & \textbf{$p_{\mathrm{rand}}$} & \textbf{Match R. $\uparrow$} & \textbf{TAcc $\uparrow$} & \textbf{Final Diversity} \\
\midrule
No pheromone & 0 & 0.05 & 21.23 & 19.86 & 0.76 \\
Full model w/o random exploration & dynamic $\beta$ & 0 & 24.63 & 26.34 & 0.52 \\
Full model & dynamic $\beta$ & 0.05 & 25.25 & 27.16 & 0.58 \\
Full model, larger random exploration & dynamic $\beta$ & 0.10 & 24.38 & 26.47 & 0.66 \\
\bottomrule
\end{tabular}
\end{table}

Overall, pheromone is the primary mechanism governing directed exploration in PhGPO, while $p_{\mathrm{rand}}$ serves as a lightweight auxiliary safeguard against early over-concentration.
This distinction also explains why the two mechanisms are complementary: pheromone reuses accumulated transition evidence, whereas random exploration preserves a small chance of escaping over-concentrated paths.

\section{Broader Impact}
\label{app:broader_impact}

This work studies long-horizon tool planning for LLM agents in benchmarked tool-use environments.
Its potential positive impact is to improve the reliability and training efficiency of tool-use agents by making successful transition patterns explicit and reusable.
More reliable tool planning may benefit applications that require multi-step interaction with software tools, such as information retrieval, code execution, data processing, and workflow automation.

At the same time, stronger tool-use agents should be deployed with care.
If connected to real external APIs, an agent may execute incorrect, unintended, or unsafe tool calls, especially when task instructions are ambiguous or tool outputs are unreliable.
Such risks can be more serious in sensitive domains involving personal data, security-relevant operations, financial actions, or medical and legal decisions.
PhGPO does not remove the need for tool-level safeguards, access control, or human oversight.

We therefore view PhGPO as a method for improving tool-use training in controlled environments, not as a complete deployment framework.
Practical deployments should use sandboxed execution, permission-gated tools, audit logs, tool allowlists, validation checks, and human confirmation for high-impact actions.
The pheromone memory should also be monitored when tool interfaces or task distributions change, since accumulated transition statistics may otherwise reflect outdated tool-use patterns.
Our experiments are conducted in benchmarked tool environments rather than real-world high-stakes deployments.



\end{document}